# PolyRoute: A Twelve-Language and Code-Mixed Benchmark for Typed-Decision Routing, with Baselines and a Leaderboard

*Dataset card, label audit, protocol, and the first thirteen leaderboard entries*

Mohammed Nayeem Khan
mnknayeem@gmail.com

Sarun Kumar K S
sarun.kumar.k.s@intel.com

Intel Corporation

**Abstract**

Typed-decision encoders answer a question about a text in one forward pass, but their behaviour outside English is rarely measured on a fixed yardstick. We release PolyRoute: 322 native-language support messages in 12 languages and 9 writing systems, 156 code-mixed or romanised messages (Hinglish, Spanglish, Arabizi), 967 optional native training rows, and a protocol for a three-way department-routing decision. Labels come from Qwen3.8-27B with a blind second pass (91% of candidates kept); an independent model, gemma4-31b, agrees on 97.2% of test items, and the largest character 5-gram similarity between any test text and any training text is 0.33. A leaderboard of 13 systems gives 61.2% for the English-root checkpoint, 77.3% for the shipped multilingual one and 96.6% for the best fine-tune; TF-IDF with logistic regression reaches 91.9%. On four withheld languages the three multilingual systems score 77%, 90% and 93%. On code-mixed text the English root (85%) is nominally ahead of the shipped multilingual checkpoint (79%; p=0.134). The split resolves differences only above about 4 points. All texts and labels are LLM-generated.

**Key findings**

1. **Dataset.** 322 native items (12 languages, 5 script groups) and 156 code-mixed items; independent-labeller agreement 97.2% [94.8–98.5], lowest in Spanish (86%); no test text has a 5-gram Jaccard similarity above 0.33 to any training text.
2. **Tracks separate, the top does not.** Shipped checkpoints 61%/63%/77%; fine-tuned multilingual systems 94%–97%; 4 systems are statistically tied with the best entry.
3. **Strong cheap baselines.** TF-IDF + logistic regression on all rows scores 91.9%; an independent LLM scores 97.2%.
4. **Weak spots are script groups, not token counts.** Fertility–accuracy correlation $\rho$=-0.34 (root) and -0.17 (shipped multilingual), n=12; the shipped checkpoint is weakest on the Indic group (63%).
5. **Reduced-exposure transfer.** On four withheld languages: shipped 77%, no per-language rows 91%, withheld-language fine-tune 90%, all-language fine-tune 93%; the base rows already contain about 30 generic rows per language.
6. **Code-mixed text is a different test.** The English-root fine-tune scores 96% on code-mixed items but 74% on native text; the English root is nominally ahead of the shipped multilingual checkpoint (85% vs 79%, p=0.134).
7. **Resolution limit and pitfalls.** Differences below about 4 points among the top systems are unresolvable at n=322; rotating option names drops the best fine-tune to 7%.

# 1 Motivation and contributions

A typed-decision encoder answers a question about a piece of text in one forward pass: it reads one hidden state per answer option and returns a distribution over the options, with no token-by-token generation. For a support desk that is attractive, because routing becomes cheap and predictable. What a deployment cannot take for granted is that such an encoder behaves in other languages as it does in the one it was developed in. The public model family we study ships an English ModernBERT-large root [Warner et al. 2024] (421M parameters) and a separate multilingual checkpoint (322M parameters, 256k-token vocabulary), with a small router that chooses between them. Which checkpoint to trust for which language, and how much task-specific fine-tuning adds, are questions that need a fixed yardstick, not ad-hoc spot checks.

This paper provides one. **PolyRoute** is a test collection of 322 native-language support messages in 12 languages and 9 writing systems, a second collection of 156 code-mixed and romanised messages, an optional pool of 967 native training rows, and an evaluation protocol with documented baselines. The question is deliberately simple (three departments), so that the quantity under test is the language and not the difficulty of the question. The format is the typed decision used by the model family, which lets every item be scored for accuracy, calibration and selective prediction with no prompt engineering.

**Intended uses.** (i) Comparing multilingual encoders or fine-tuning recipes on routing-style decisions; (ii) auditing a system for language gaps before deployment; (iii) studying cross-lingual transfer through a ready-made withheld-language variant; (iv) testing robustness to code-mixed text. **Not intended** as a measurement of real-world traffic: all texts are LLM-written (Section 13).

## 1.1 Contributions

1. A native-language test split with per-language label statistics and an independent label audit (97.2% agreement; lowest in Spanish at 86%).
2. A code-mixed split (Hinglish, Spanglish, Arabizi) disjoint from every training file.
3. Five fine-tuning sets with nested native content, one of them withholding four languages, together with an explicit account of what each contains — including a block of 399 generic multilingual rows that even the 'withheld' set still contains.
4. A leaderboard of 13 systems with intervals, rank-stability analysis and the smallest difference the split can resolve (55 of 78 system pairs are separated at $p<0.01$).
5. Analyses a user will want to reproduce: per language and script, tokenizer fertility, the shipped router, learning curve, transfer to withheld languages, calibration, stress conditions and an error taxonomy with real examples.

## 1.2 Where it sits among multilingual benchmarks

Established suites such as XNLI [Conneau et al. 2018] and XTREME [Hu et al. 2020] test cross-lingual transfer on standard tasks, built on multilingual pre-trained encoders [Pires et al. 2019; Conneau et al. 2020]; English intent-detection sets such as Banking77 [Casanueva et al. 2020] are the closest in task shape; MASSIVE [FitzGerald et al. 2022] covers intent classification in 51 languages with human translation; GLUECoS [Khanuja et al. 2020] targets code-switching. PolyRoute is narrower. It fixes one decision type and one question text and is shaped by the typed-decision interface, so that models which score options through placeholder tokens can be evaluated without a generation step. The price of that narrowness is that conclusions about other tasks do not follow; the benefit is a compact split with documented provenance. We follow datasheet [Gebru et al. 2018] and data-statement [Bender and Friedman 2018] practice in Sections 4 and 13, and the advice on statistical power in NLP evaluation [Card et al. 2020; Berg-Kirkpatrick et al. 2012] in Section 7.

**Table 1.** The benchmark at a glance.

| Item | Value |
|---|---|
| Task | single-label department routing (billing / technical / other) as one typed decision, one forward pass |
| Native-language test split | 322 items, 12 languages, 9 writing systems (Japanese kana-plus-Han counted separately; 5 coarser script groups); billing 118, technical 116, other 88 |
| Code-mixed test split | 156 items: Hinglish 58, Spanglish 47, Arabizi 51 |
| Native training rows (optional) | 967 rows from a separate generation run, to be added to 2,818 base rows |
| Labels | written by Qwen3.8-27B and kept only if a blind second pass by the same model agrees; audited by gemma4-31b |
| Independent agreement | 97.2% [94.8–98.5] on the native split (313 consensus items) |
| Leakage | largest character 5-gram Jaccard similarity of any test text to any training text: 0.33 (median 0.08); exact duplicates: 0 |

| Item | Value |
|---|---|
| Primary metric | accuracy with an exact (Wilson) 95% interval; secondary: language-macro accuracy, worst language, macro-F1, NLL, ECE |
| Leaderboard | 13 systems: nine Laya-based, three TF-IDF + logistic regression, one independent LLM |

## 2 Task definition

### 2.1 The typed decision

Every item is a pair (text, question). The question is of type *choice*: an instruction ("Which department should handle this?") and an ordered set of named options with a one-line description each: `billing` (invoices, payments, refunds), `technical` (bugs, outages, system errors) and `other` (everything else). Instruction and options are in English in every item whatever the language of the message, mirroring a deployment in which the operator writes the schema once and messages arrive in many languages. The gold answer is one option.

### 2.2 How the model under test reads an item

The Laya family packs question and text into one sequence: a classification token, the question and instruction, a separator, a placeholder token in front of each option name and description, a separator, and the message, truncated to 512 tokens in total (192 for the question part). The encoder runs once; a two-layer option head reads the hidden state at each placeholder and emits one score per option, and a softmax over the scores gives the confidence. The root checkpoint has 28 encoder layers of width 1024; the multilingual checkpoint has 22 layers of width 768, with 61% of its parameters in the embedding table. Neither the data nor the metrics depend on this architecture: any system that returns a distribution over the three options can be submitted.

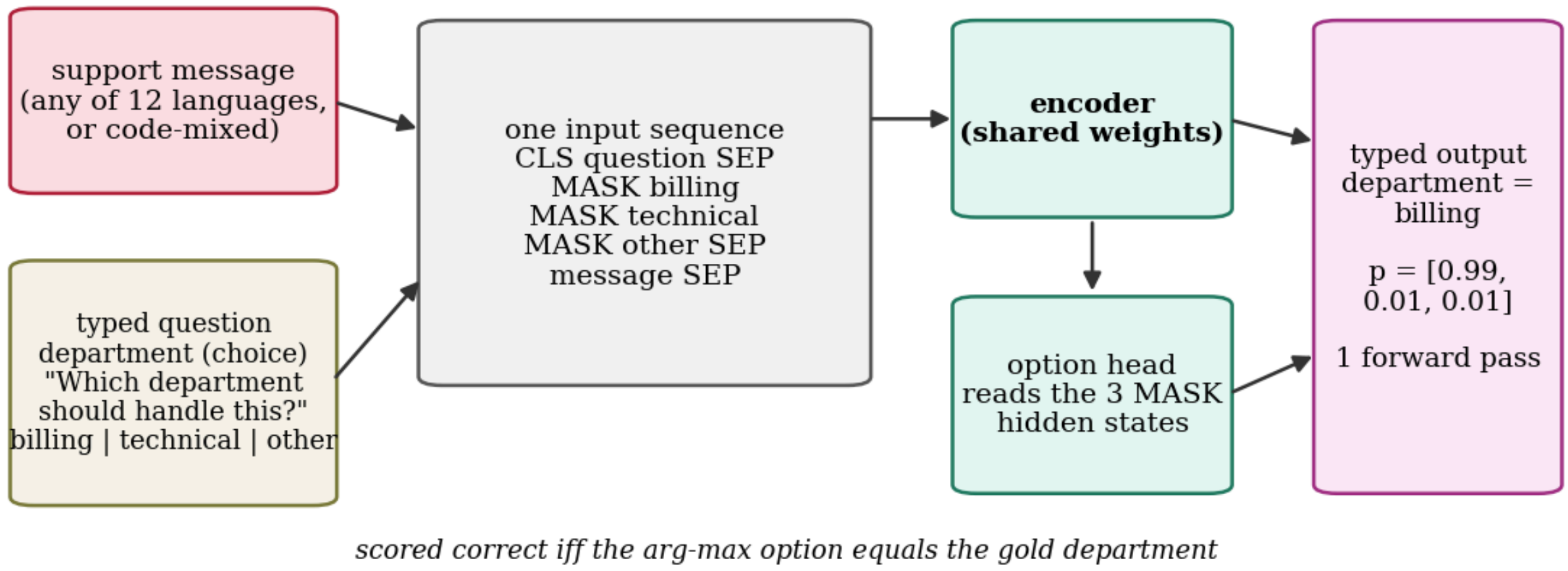

**Figure 1.** The benchmark item. A message and a typed question are packed into one sequence with a placeholder in front of each option; the model reads the hidden state at every placeholder and returns one probability per option (the numbers shown are the actual output of the best fine-tuned system on the first test item).

### 2.3 Why routing

Department routing is topical: the cue is lexical or semantic (a refund, an outage) and not syntactic (negation, word order). That keeps the items answerable in every language with simple vocabulary, so an error is unlikely to be explained by the question being too hard for a native speaker. The price is a high ceiling: tuned systems score above 95% (Section 8), so the benchmark discriminates mainly among systems not tuned to it, and Section 7 quantifies how little room is left at the top.

## 3 Construction of the data

### 3.1 Generation

For each language and each department the teacher model (Qwen3.8-27B, sampling temperature 0.9) is asked for up to ten diverse, realistic support messages in that language only, in its native script, with no English words except product names, each of which has the requested department as its correct answer (label-conditioned self-generation in the style of [Wang et al. 2022]); length, style and vocabulary are varied through a per-call variation seed. Across the native test split and the native training rows this produced 1,420 candidate texts.

### 3.2 Blind filtering

The candidates of each call are shown to the same model again without their intended labels, together with the three options, and it picks one option per text. A text survives only if the blind choice equals the intended label: 1,289 of 1,420 candidates (90.8%) did. Retention differs by department (billing 95%, technical 97%, other 80%) and by language, from 86% for Hindi to 97% for Turkish (Figure 2). The filter removes some genuinely ambiguous messages, but it is a self-consistency check and not a human one, which is why Section 5 adds an audit by a different model family.

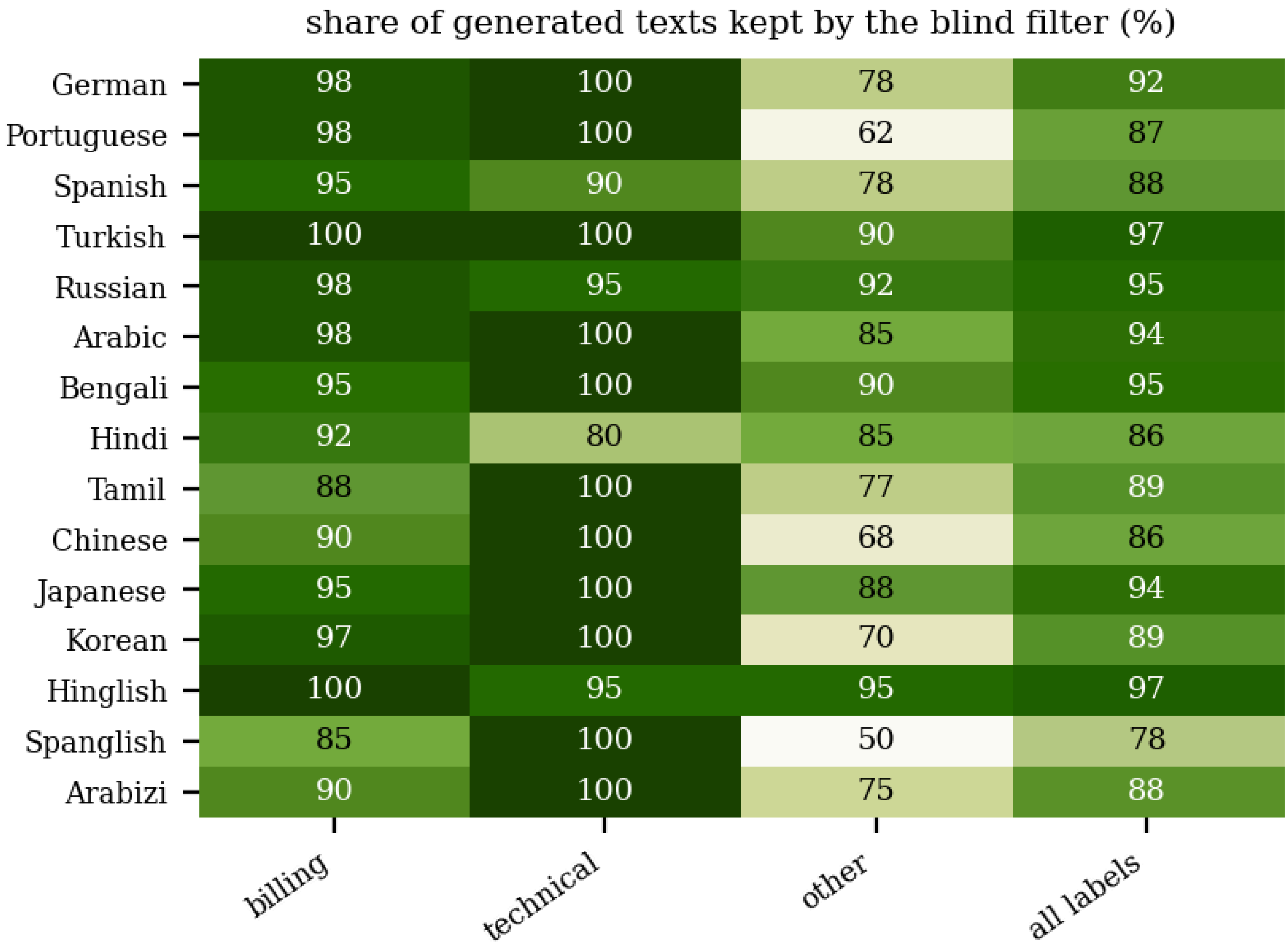

**Figure 2.** Filter retention for every language-by-label cell. Low cells flag where the teacher's intended label and its blind reading diverge, which is a first, language-level indicator of label difficulty.

### 3.3 Separation of files

The test split (322 items) and the training rows (967 rows) come from two generation runs with different variation seeds, and during each run any text identical to a text in another split file is dropped. The code-mixed split was generated the same way from three prompts: Hindi in Latin letters freely mixed with English; Spanish mixed with English inside a sentence; and Arabic written in Latin letters and digits, mixed with English. The prompts ask for chat-like text and not for textbook code-switching. Of 180 candidates, 158 passed the filter and 156 are in the released file.

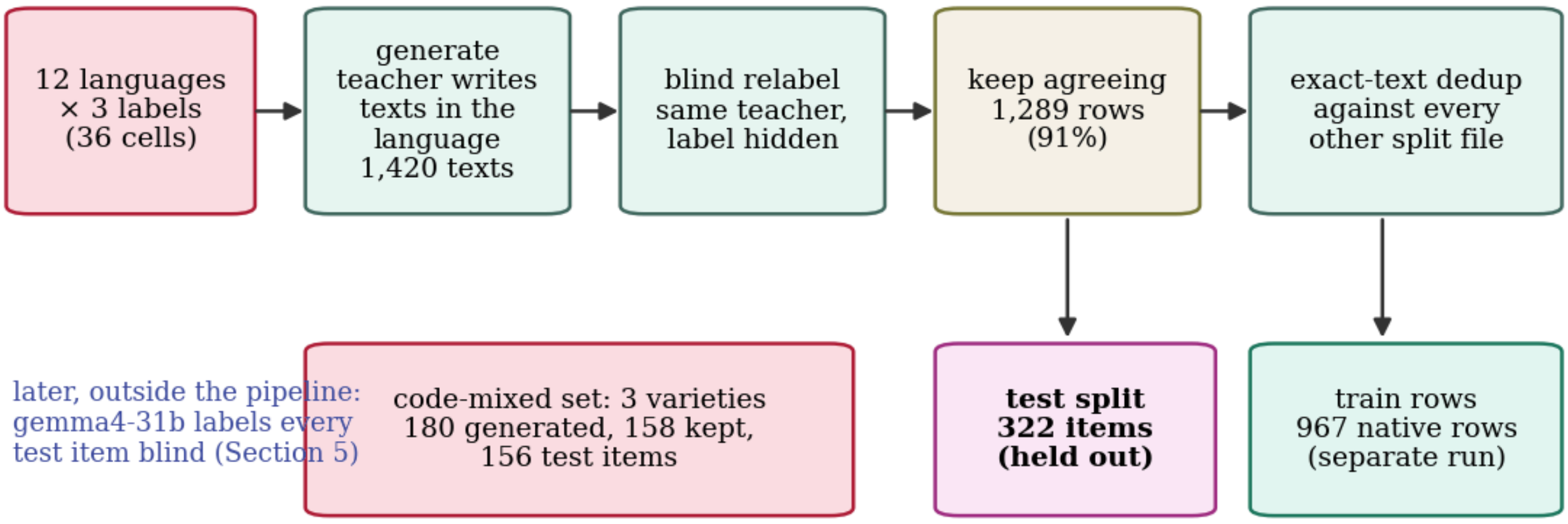

**Figure 3.** How the benchmark files were produced and what survived at each step (counts are the real generation log; the test split and the native training rows come from separate generation runs with different variation seeds).

## 4 Dataset card

The native split holds 322 messages, between 21 and 30 per language, with 118 billing, 116 technical and 88 other items. The residual class is the smallest because it loses most candidates in the filter (Figure 2), and the balance per language is uneven: Portuguese has 2 other item(s), Korean has 1 other item(s), so for those languages accuracy mostly reflects billing and technical messages. Table 2 is the data card; messages are short (median 108 characters), and for every language the text is overwhelmingly in the expected writing system (mean script purity 98%; the lowest is Korean at 95%).

**Table 2.** Per-language data card for the native split. Tier is a coarse, author-assigned grouping by typical pre-training availability (not measured here). Script purity is the share of alphabetic characters in the expected writing system. Tokens are counted without special tokens by each checkpoint's own tokenizer.

| Language | Script | Tier | Test n | bill/tech/other | Train rows | Median chars | Median tokens root / multi. | Script purity | Indep. agreement |
|---|---|---|---|---|---|---|---|---|---|
| German | Latin | high | 27 | 10/10/7 | 83 | 172 | 55 / 40 | 100% | 100% |
| Portuguese | Latin | high | 22 | 10/10/2 | 82 | 88 | 28 / 22 | 100% | 100% |
| Spanish | Latin | high | 28 | 10/10/8 | 77 | 160 | 46 / 32 | 100% | 86% |
| Turkish | Latin | mid | 28 | 10/10/8 | 85 | 117 | 54 / 32 | 100% | 100% |
| Russian | Cyrillic | high | 28 | 9/9/10 | 86 | 122 | 58 / 34 | 97% | 96% |
| Arabic | Arabic | mid | 28 | 10/10/8 | 85 | 136 | 94 / 46 | 99% | 100% |
| Bengali | Bengali | low | 26 | 10/7/9 | 81 | 116 | 160 / 65 | 99% | 96% |
| Hindi | Devanagari | mid | 27 | 9/10/8 | 76 | 110 | 103 / 39 | 95% | 96% |
| Tamil | Tamil | low | 28 | 10/10/8 | 74 | 103 | 148 / 46 | 98% | 96% |
| Chinese | Han | high | 29 | 10/10/9 | 74 | 43 | 60 / 29 | 98% | 100% |
| Japanese | Han + kana | high | 30 | 10/10/10 | 83 | 54 | 56 / 29 | 99% | 100% |
| Korean | Hangul | mid | 21 | 10/10/1 | 81 | 37 | 59 / 26 | 95% | 95% |
| all | – | – | 322 | 118/116/88 | 967 | 108 | 62 / 34 | 98% | 97% |

### 4.1 Length and tokenization

Character length varies by a factor of 4.6 between languages (the shortest messages, in characters, are Korean, Chinese, Japanese), but token counts matter more to an encoder. The English-root tokenizer needs 2.0× as many tokens as the multilingual one on average, and up to 3.1× for Tamil (Table 2). The longest message is 211 root tokens, so truncation at 512 tokens never occurs: inflation changes what the model sees, not how much of it.

### 4.2 The code-mixed split

**Table 3.** The code-mixed split. 'Latin share' is the fraction of alphabetic characters in Latin script, which is the defining property of romanised text.

| Variety | n | bill/tech/other | Median chars | Latin share | Generated → kept |
|---|---|---|---|---|---|
| Hinglish | 58 | 20/19/19 | 86 | 100% | 60 → 58 |

| Variety | n | bill/tech/other | Median chars | Latin share | Generated → kept |
|---|---|---|---|---|---|
| Spanglish | 47 | 17/20/10 | 152 | 100% | 60 → 47 |
| Arabizi | 51 | 18/20/13 | 155 | 100% | 60 → 53 |

The three varieties are written almost entirely in Latin letters (Hinglish 100%, Spanglish 100%, Arabizi 100%), including the Arabic and Hindi ones, which is a plausible source of difficulty for a model whose Hindi and Arabic knowledge is attached to native-script tokens. There are 156 items; interval widths at that size are wide (Section 9.6).

### 4.3 A block of generic multilingual rows in the base training set

The 2,818 base fine-tuning rows, shared by every training track, include 399 department-routing rows produced by an earlier generic prompt that asked for texts "in a different language from a list of twelve". They are not labelled by language; by dominant script they split as Latin 138, Han + kana 34, Arabic 34, Cyrillic 33, Han 33, Bengali 33, Hangul 32, Tamil 31, Devanagari 31, i.e. roughly 30–35 rows for each non-Latin language. They are disjoint from the test split by the filters of Section 3, but they mean that **no training track is free of exposure to the twelve languages**. The track names below say 'no per-language rows' or 'withheld', never 'zero-shot'.

### 4.4 Language fidelity of the generated text

The script-purity column of Table 2 is a coarse check. Two cheap, stricter checks find generation artefacts that we left in place. First, 3 of the 217 non-Latin-script test messages contain at least one letter from a writing system that is neither the expected one nor Latin (Bengali 1, Hindi 1, Japanese 1). Second, a stop-word heuristic flags 2 of the 105 Latin-script test messages as English text filed under Turkish, which script purity cannot detect. These items are valid routing questions with correct departments, but they are not clean samples of the named language, and a user who needs strictly monolingual text should filter them (the indices are reproducible from the released file).

## 5 Label quality and leakage

### 5.1 Independent labelling

Because the gold labels come from one model family, we asked a different family (gemma4-31b) to label every test item blind, with the same prompt format as the generation-time filter. It agrees with the gold label on 313 of 322 items (97.2% [94.8–98.5]). Agreement is high in every language (Figure 4): the lowest per-language point estimate is Spanish (86%, 4 of 28 items disputed) and 6 of 12 languages are at 100%. At about 27 items per language the intervals are wide, and only Spanish is distinguishable from the pooled value by eye; we make no per-language claims about label quality.

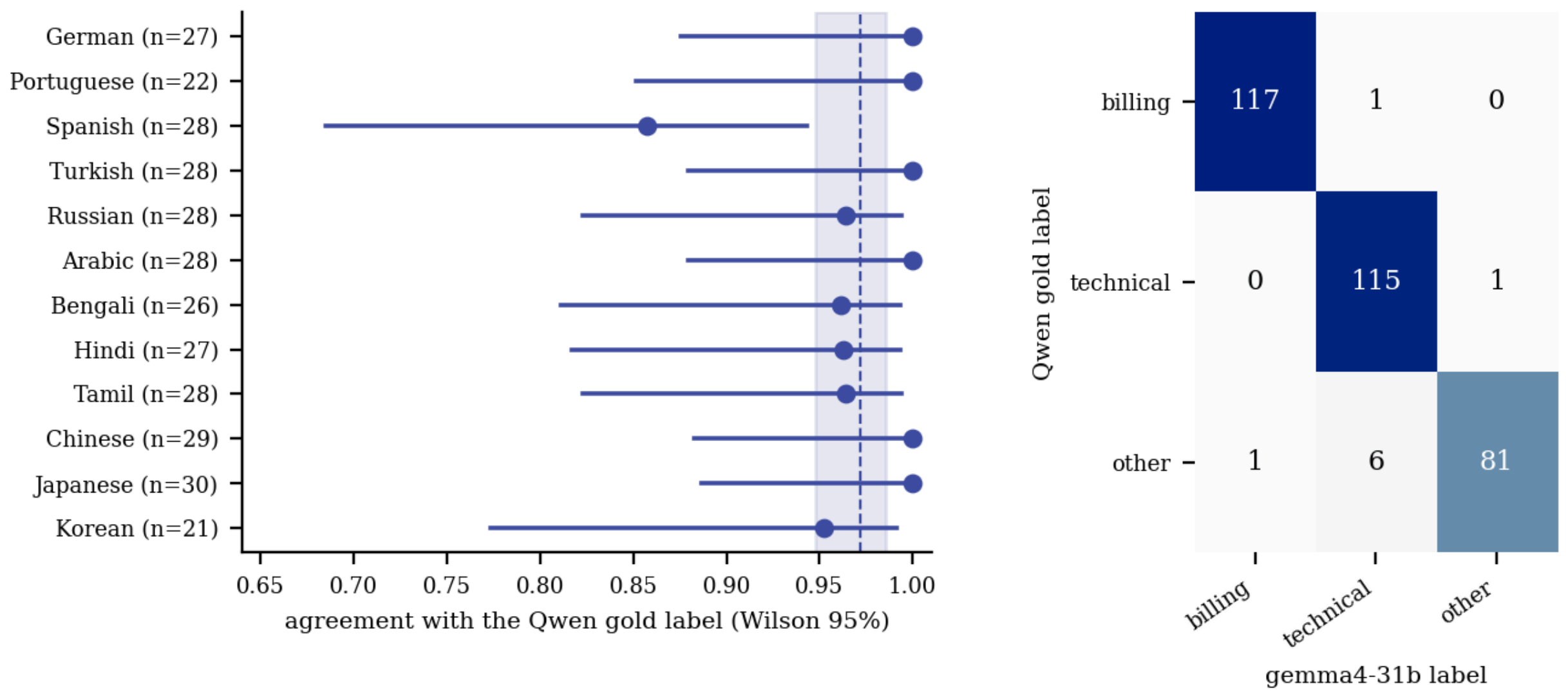

**Figure 4.** Label quality measured by an independent model family. Left: per-language agreement of gemma4-31b with the gold label (band: interval for the pooled split; dashed line: pooled agreement). Right: gold-versus-independent confusion counts; the off-diagonal cells are the disputed items.

The disagreements are not spread evenly across departments: agreement is 99.2% for billing (n=118), 99.1% for technical (n=116), 92.0% for other (n=88). The most frequent cell is gold 'other' labelled 'technical' by the independent model (6 items), consistent with the residual class being the least well delimited: a message that mentions a payment problem caused by an outage can legitimately be argued either way. We define the **consensus subset** as the 313 items on which both labellers agree, and report it as a secondary view in the leaderboard (the confusion counts are in the right panel of Figure 4).

### 5.2 Does label noise explain the model errors?

It can be bounded directly. The best fine-tuned system (ML fine-tune, all native rows) scores 97.1% on the consensus subset against 77.8% on the 9 disputed items; the shipped multilingual checkpoint scores 79.2% against 11.1%. Systems are less accurate on disputed items, as expected when gold labels are noisy there, but the 9 disputed items could account for at most 9 of the 125 errors of the root checkpoint, so the gap between weak and strong systems is real and not an artefact of the labels. For strong systems the picture is different: 2 of the 11 errors of the best fine-tuned system fall on disputed items (a minority), so label noise is a visible but not dominant share of what remains.

### 5.3 Leakage audit

For every test text we computed the maximum Jaccard similarity of its character 5-gram set to any text in the pool of all 3,785 training rows (and, for the native split, to the rest of the test split). Table 4 and Figure 5 summarise the result: no native test text exceeds 0.33 against the training pool, no code-mixed text exceeds 0.20, and the most similar pair of native test texts has similarity 0.17; there are 0 exact duplicates with training rows and 0 inside the test split. Several items in the same language share topical vocabulary (refund, invoice, crash), which probably accounts for most of the similarity mass near 0.05–0.15, but nothing resembling a character-level near-duplicate (a paraphrase would not be caught by this measure).

**Table 4.** Leakage statistics: maximum character 5-gram Jaccard similarity of each text to any text in the comparison pool, and the number of exact duplicates (after lower-casing and whitespace normalisation).

| Comparison | Texts | Median | 95th pct | Max | ≥0.3 | ≥0.5 | Exact dup. |
|---|---|---|---|---|---|---|---|
| native test vs. training rows | 322 | 0.076 | 0.194 | 0.328 | 3 | 0 | 0 |
| code-mixed test vs. training rows | 156 | 0.070 | 0.146 | 0.201 | 0 | 0 | 0 |
| native test vs. rest of test | 322 | 0.045 | 0.104 | 0.170 | 0 | 0 | 0 |

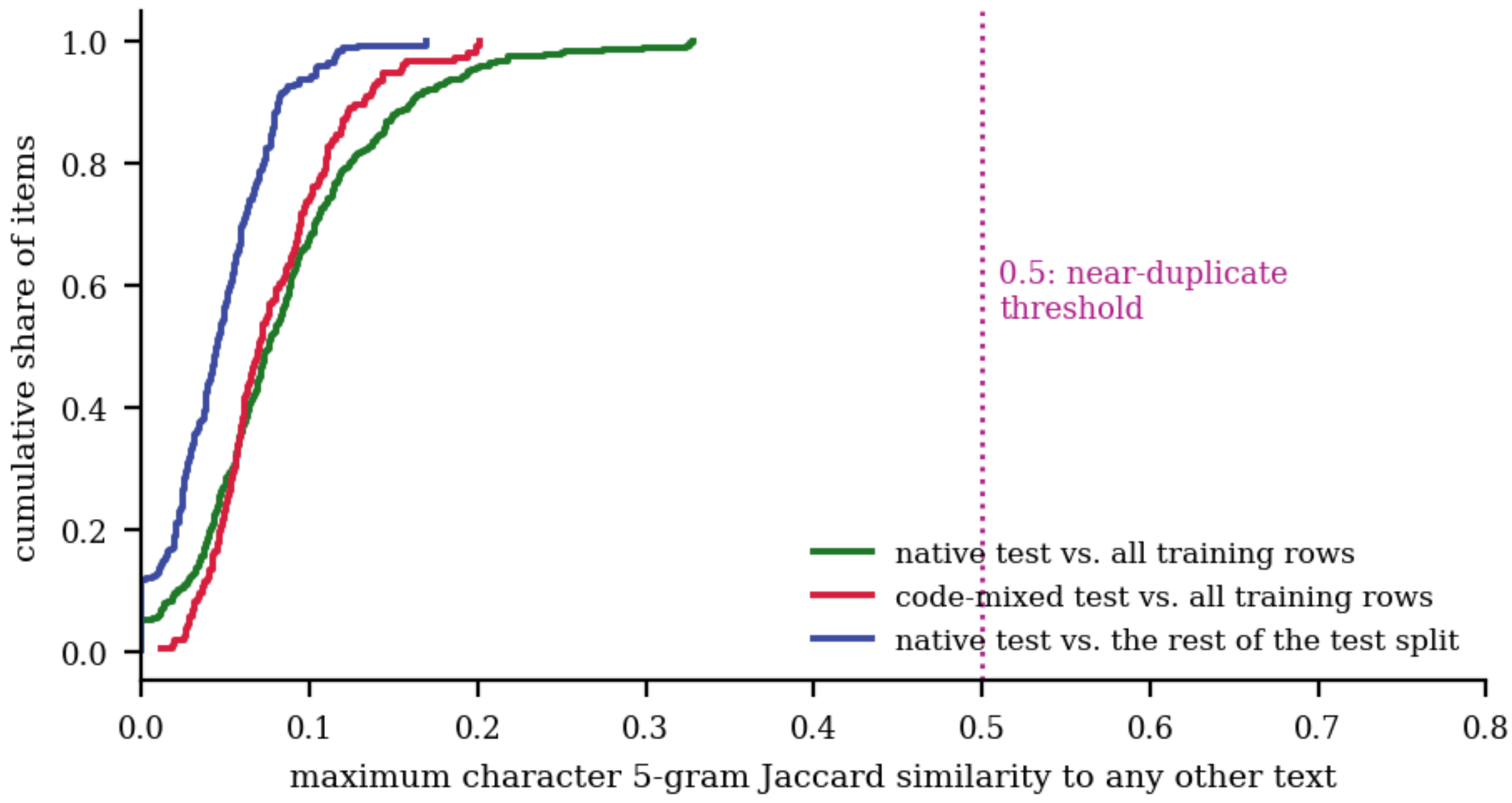

**Figure 5.** Leakage audit as empirical cumulative distributions. Every curve reaches 1 well to the left of the 0.5 line, so no test text has a close relative in the training rows or elsewhere in the test split.

## 6 Splits, tracks and evaluation protocol

### 6.1 Splits

There are two held-out evaluation splits, never to be used for training or model selection: the native split (322 items) and the code-mixed split (156 items). Training material is optional: the base rows and the 967 native rows are released so that results are comparable, and Table 5 lists the five nested sets we trained on. The sets differ only in the amount (and language coverage) of dedicated native rows, so differences between models trained from the same checkpoint are attributable to that quantity.

**Table 5.** The fine-tuning sets. All start from the same 2,818 base rows (of which 399 are generic multilingual); the multilingual encoder is fine-tuned with Laya's own trainer (4 epochs, option shuffling, the same settings for every set). The English-root fine-tune uses the base rows only.

| Track | Model | Total rows | Dedicated native rows | of which Tamil/Bengali/Korean/Arabic |
|---|---|---|---|---|
| B | English-root fine-tune | 2,818 | 0 | 0 |
| B | ML fine-tune, no per-language rows | 2,818 | 0 | 0 |
| C | ML fine-tune, 25% native rows | 3,059 | 241 | 83 |
| C | ML fine-tune, 50% native rows | 3,301 | 483 | 162 |
| B | ML fine-tune, 4 languages withheld | 3,464 | 646 | 0 |
| C | ML fine-tune, all native rows | 3,785 | 967 | 321 |

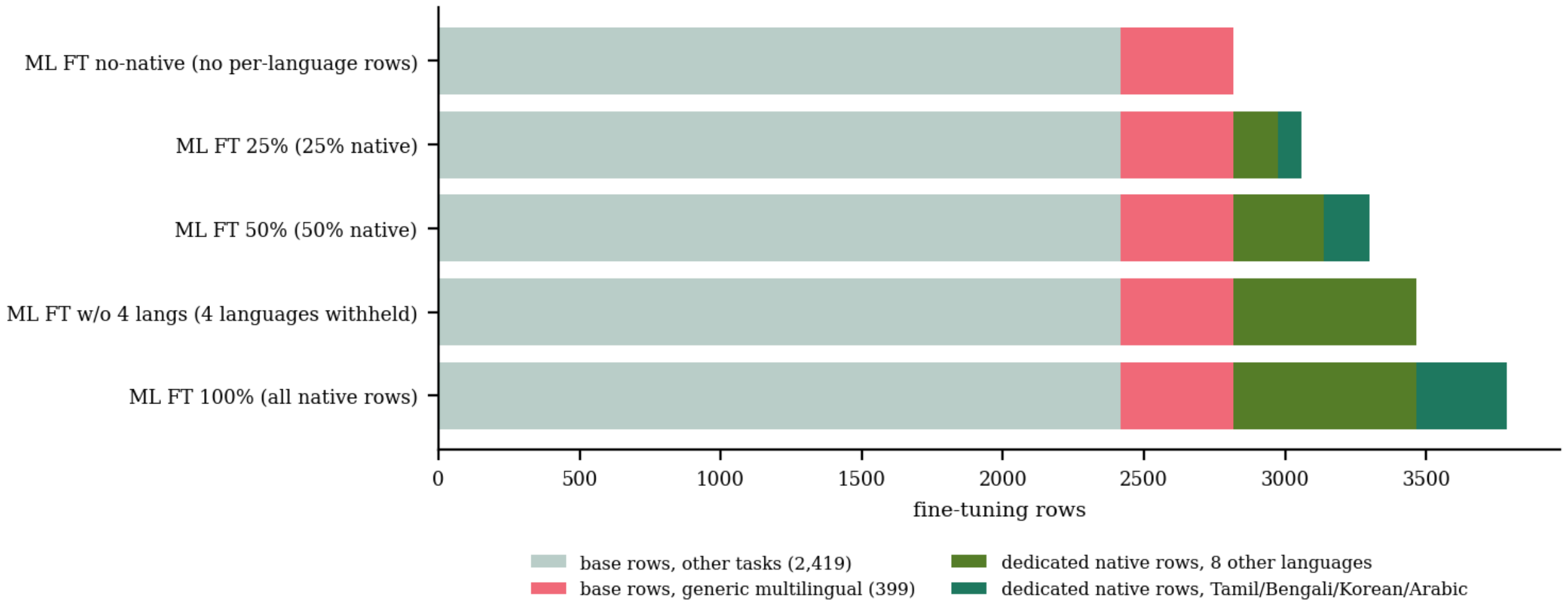

**Figure 6.** Composition of the five fine-tuning sets behind the benchmark's training tracks. All start from the same base rows; the generic multilingual block is present in every set, including the 'withheld' one, which is why that set measures reduced exposure rather than zero exposure.

## 6.2 Tracks and rules

Entries are grouped in tracks (Table 6) so that a system fine-tuned with native rows is not ranked against a checkpoint used as released without saying so. For the withheld-language variant, the four languages (Tamil, Bengali, Korean, Arabic) were chosen so that each uses a writing system that none of the eight remaining languages uses (Tamil script, Bengali script, Hangul, Arabic script); a model cannot therefore transfer script-specific vocabulary from the languages that stay in training.

**Table 6.** Tracks and submission rules.

| Track | What may be used | Typical entrant |
|---|---|---|
| A: as released | no fine-tuning on any PolyRoute file; the checkpoint is evaluated as published | root, typed-decisions and multilingual checkpoints |
| B: no dedicated native rows | task rows without per-language native data (the generic multilingual block may be present and must be declared); or native rows with the four withheld languages removed | English-root fine-tune; ML fine-tune without per-language rows; withheld-language variant |
| C: native-row tuning | all or a declared fraction of the native training rows | ML fine-tunes with 25%, 50% and 100% of the rows |
| Reference | classical or prompted systems that do not use the typed-decision interface; reported for context, not ranked against tracks A–C | TF-IDF + LR; independent LLM |

## 6.3 Evaluation protocol

1. Present each item with the fixed question text and the options in the canonical order (billing, technical, other); do not rename options, because Section 10 shows how strongly the models follow option names.
2. Obtain one probability per option; the prediction is the arg-max, the confidence is its probability.

3. Report accuracy with a Wilson interval on the native split and, separately, on the code-mixed split; also report language-macro accuracy and the worst language.
4. For pairwise claims use an exact McNemar test on the shared items and state the number of comparisons made.
5. Never tune on the test items; if a threshold is calibrated, calibrate it on a held-out part of the training rows.
6. Declare the exposure of the fine-tuning set to the twelve languages (including generic multilingual rows) and the number of seeds.

# 7 Metrics and statistical resolution

## 7.1 Metrics

**Table 7.** Metrics reported on the leaderboard.

| Metric | Definition | Why |
|---|---|---|
| Accuracy | share of items whose arg-max option equals the gold option; Wilson 95% interval | primary; exact interval is valid at the sample sizes here |
| Language-macro accuracy | mean of the twelve per-language accuracies; stratified bootstrap interval (resampling within language) | guards against one language dominating |
| Worst language | lowest per-language accuracy and its language | the number a multilingual deployment is limited by (about 27 items, so indicative) |
| Macro-F1 | unweighted mean of per-department F1 | classes are not balanced |
| NLL | mean negative log-probability of the gold option | quality of the whole distribution |
| ECE | expected calibration error, 10 equal-width confidence bins | does confidence mean what it says |
| Selective accuracy | accuracy on the most confident 80% of items; area under the risk-coverage curve | usable when a system may abstain |
| Consensus accuracy | accuracy on items the independent labeller also agrees with | a lower-noise view of the same split |

## 7.2 How large a difference can the split resolve?

A benchmark with 322 items cannot rank systems whose accuracies are close. For a paired comparison in which two systems disagree on a fraction d of the items, the smallest difference detectable at $\alpha=0.01$ with 80% power is about $3.42\sqrt{(d/n)}$ (a normal approximation to McNemar's test). Table 8 evaluates this for the sizes that occur in this benchmark; Figure 7 shows the curve and places every pair of leaderboard systems against it. The top five fine-tuned systems disagree on 3.6% of the items on average, so on the native split the resolvable difference among them is 3.6 points — larger than the largest gap between them (2.8 points). For a single language (about 26 items) even a 25%-discordant pair needs a 34-point gap, which is why we do not rank systems per language. We use the single threshold $p<0.01$ without correction for the 78 leaderboard pairs; with a Bonferroni threshold of 0.01/78, 48 of the 55 separated pairs remain separated. Anyone claiming a gain of a few points should repeat training with several seeds and use the paired bootstrap over items.

**Table 8.** Smallest detectable accuracy difference (points) at $\alpha=0.01$ and 80% power for paired comparisons, by test size and discordance rate.

| Test items | 3% discordant | 6% | 12% | 25% |
|---|---|---|---|---|
| 322 (native split) | 3.3 | 4.7 | 6.6 | 9.5 |
| 156 (code-mixed split) | 4.7 | 6.7 | 9.5 | 13.7 |
| 26 (one language) | 11.6 | 16.4 | 23.2 | 33.5 |
| 64 (one script group) | 7.4 | 10.5 | 14.8 | 21.4 |

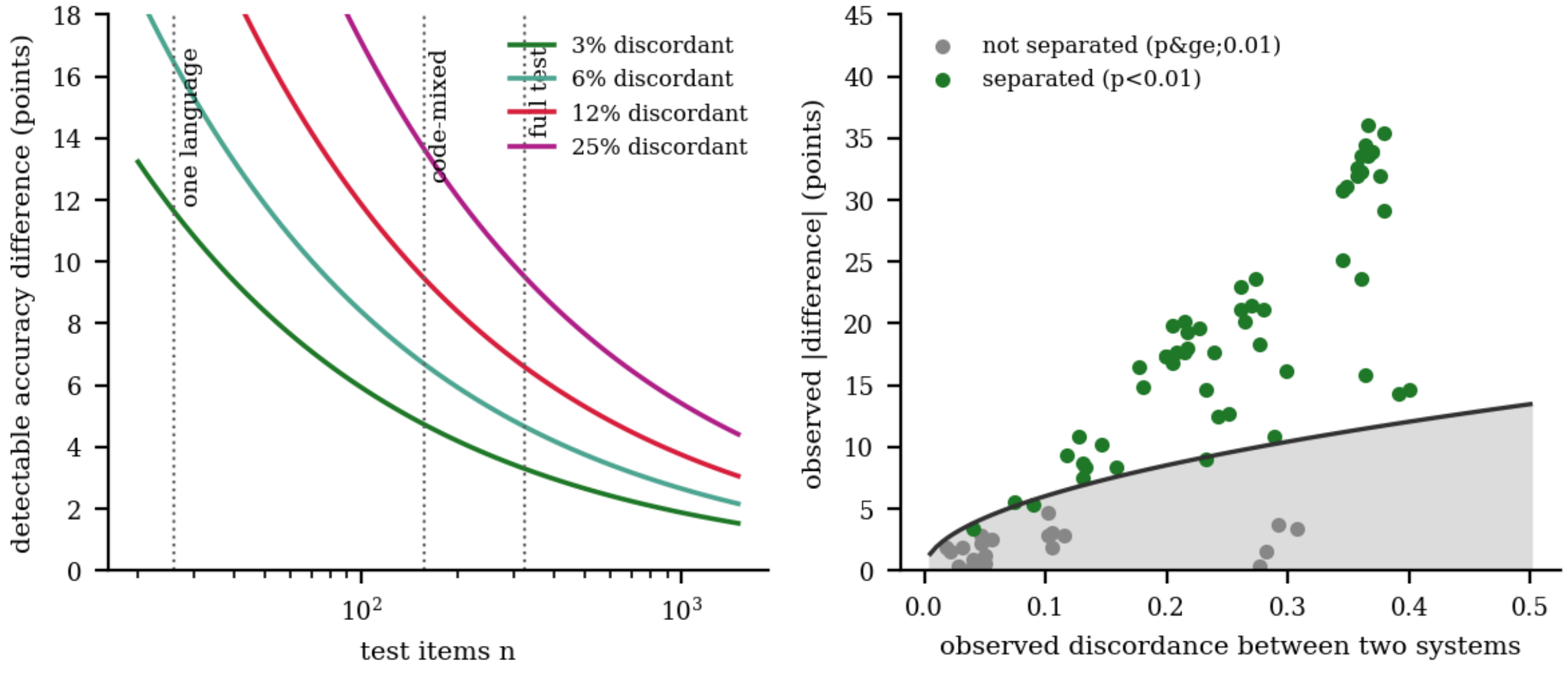

**Figure 7.** What the test split can resolve. Left: smallest accuracy difference detectable at α=0.01 and 80% power as a function of test size, for several discordance rates (the share of items on which two systems differ). Right: every pair of leaderboard systems at n=322; the shaded region lies below the detection frontier.

## 8 Systems and leaderboard

### 8.1 Systems

Thirteen systems are on the first leaderboard (Table 9). Nine are Laya-based and fall into the three tracks of Section 6; the other four are references that show how much of the benchmark can be solved without the typed-decision encoder: three TF-IDF + logistic-regression models and an independent LLM used as a classifier.

**Table 9.** The systems on the leaderboard. Latency is the measured time per item on one RTX 4090 at batch size 16 where it was recorded.

| System | Track | What it is | Params | ms / item |
|---|---|---|---|---|
| Laya root (English ModernBERT-large) | A | released English root: ModernBERT-large encoder and two-layer option head | 421M | 1.6 |
| Laya typed-decisions checkpoint | A | released fine-tune of the English root on the authors' typed-decision data | – | 1.6 |
| Laya-multilingual (shipped) | A | released multilingual checkpoint (22 layers, width 768, 256k vocabulary) | 322M | 1.2 |
| English-root fine-tune | B | English root fine-tuned on the base rows only | as root | 3.1 |
| ML fine-tune, no per-language rows | B | multilingual checkpoint fine-tuned on the base rows only | as shipped | 2.1 |
| ML fine-tune, 4 languages withheld | B | base rows + native rows of eight languages | as shipped | – |
| ML fine-tune, 25% native rows / 50% native rows | C | base rows + a random 25% / 50% of the native rows | as shipped | – / – |
| ML fine-tune, all native rows | C | base rows + all native rows (three seeds) | as shipped | 2.0 |
| TF-IDF + LR (three variants) | ref | word 1–2-grams plus character 2–5-grams, logistic regression (C=10); trained on the department rows of the matching set: 740, 1,386, 1,707 rows | – | – |
| gemma4-31b as classifier | ref | different model family from the teacher, prompted to choose one option blind (the label-audit run) | – | network-bound |

### 8.2 Leaderboard

Tables 10 and 11 give the full results; Figure 8 draws the primary metric. Four observations follow.

1. **Tracks separate cleanly; the top does not.** The shipped checkpoints (track A) score 61.2%, 62.7% and 77.3%; every fine-tuned multilingual system (tracks B and C) scores between 93.8% and 96.6%. The highest accuracy is gemma4-31b as classifier at 97.2%, and 4 other systems cannot be separated from it at $p<0.01$ (ML FT w/o 4 langs, ML FT 25%, ML FT 50%, ML FT 100%).

2. **A bag of n-grams is a serious baseline.** TF-IDF + LR trained on all rows scores 91.9%, +4.7 points (p=0.014) from the best fine-tuned model; trained on the base rows only it scores 77.0%, level with the shipped multilingual checkpoint (-0.3 points (p=1.000) for TF-IDF relative to it). Department routing is lexically easy, and any claim about a multilingual encoder on this task has to be read against that.
3. **An independent LLM, used as a classifier, is a strong reference.** gemma4-31b agrees with the gold labels on 97.2%, +0.6 points (p=0.804) relative to the best fine-tuned model. It was not trained on these labels, so this is also a ceiling of sorts for the label audit; it costs a hosted model and a network round trip per item, against 2.0 ms for the fine-tuned encoder.
4. **Multilingual fine-tunes are better calibrated.** ECE falls from 0.058 (root) and 0.078 (shipped multilingual) to 0.011 for the best fine-tune, whereas the English-root fine-tune has 0.166 (Section 10).

**Table 10.** Leaderboard, part 1: accuracy and probabilistic quality on the 322-item native split. Track A: shipped checkpoints; B: fine-tuned without dedicated per-language rows; C: fine-tuned with native rows; ref: reference systems outside the typed-decision interface. Macro-F1 needs predicted labels, which the TF-IDF runs did not store.

| System | Track | Train rows | Accuracy [95% CI] | Macro-F1 | NLL | ECE |
|---|---|---|---|---|---|---|
| Laya root (English ModernBERT-large) | A | none | 61.2% [55.8–66.3] | 0.563 | 0.806 | 0.058 |
| Laya typed-decisions checkpoint | A | none | 62.7% [57.3–67.8] | 0.632 | 0.872 | 0.124 |
| Laya-multilingual (shipped) | A | none | 77.3% [72.4–81.6] | 0.735 | 0.653 | 0.078 |
| English-root fine-tune | B | 2,818 | 73.6% [68.5–78.1] | 0.728 | 0.844 | 0.166 |
| ML fine-tune, no per-language rows | B | 2,818 | 93.8% [90.6–95.9] | 0.934 | 0.204 | 0.038 |
| ML fine-tune, 4 languages withheld | B | 3,464 | 95.0% [92.1–96.9] | 0.948 | 0.175 | 0.020 |
| ML fine-tune, 25% native rows | C | 3,059 | 94.7% [91.7–96.7] | 0.945 | 0.198 | 0.019 |
| ML fine-tune, 50% native rows | C | 3,301 | 94.7% [91.7–96.7] | 0.945 | 0.179 | 0.022 |
| ML fine-tune, all native rows | C | 3,785 | 96.6% [94.0–98.1] | 0.965 | 0.125 | 0.011 |
| TF-IDF + LR, base rows only | ref | 740 | 77.0% [72.1–81.3] | – | – | – |
| TF-IDF + LR, 4 languages withheld | ref | 1,386 | 86.3% [82.2–89.7] | – | – | – |
| TF-IDF + LR, all rows | ref | 1,707 | 91.9% [88.4–94.4] | – | – | – |
| gemma4-31b as classifier | ref | none (prompted) | 97.2% [94.8–98.5] | 0.970 | – | – |

**Table 11.** Leaderboard, part 2: secondary views. Language-macro accuracy averages the twelve per-language accuracies (stratified bootstrap interval). The withheld-language column covers Tamil, Bengali, Korean and Arabic. The consensus subset keeps items on which the independent labeller agrees with the gold label.

| System | Language-macro acc. [95% CI] | Worst language | Withheld four (n=103) | Consensus (n=313) | Code-mixed (n=156) |
|---|---|---|---|---|---|
| Laya root (English ModernBERT-large) | 61.2% [56.4–66.3] | Arabic 32% | 39.8% | 62.6% | 84.6% [78.1–89.4] |
| Laya typed-decisions checkpoint | 62.1% [57.3–67.4] | Tamil 29% | 34.0% | 63.6% | – |
| Laya-multilingual (shipped) | 77.7% [72.6–82.1] | Hindi 56% | 76.7% | 79.2% | 79.5% [72.5–85.1] |
| English-root fine-tune | 73.1% [69.0–77.6] | Bengali 35% | 44.7% | 73.8% | 96.2% [91.9–98.2] |
| ML fine-tune, no per-language rows | 93.7% [91.3–96.2] | Spanish 79% | 91.3% | 96.2% | 94.9% [90.2–97.4] |
| ML fine-tune, 4 languages withheld | 94.9% [92.5–97.2] | Tamil 86% | 90.3% | 96.5% | 96.2% [91.9–98.2] |
| ML fine-tune, 25% native rows | 94.6% [92.3–96.8] | Bengali 85% | 91.3% | 95.8% | 93.6% [88.6–96.5] |
| ML fine-tune, 50% native rows | 94.6% [92.1–96.9] | Spanish 86% | 92.2% | 96.2% | 94.9% [90.2–97.4] |
| ML fine-tune, all native rows | 96.5% [94.6–98.3] | Bengali 88% | 93.2% | 97.1% | 96.2% [91.9–98.2] |
| TF-IDF + LR, base rows only | 77.3% [72.3–81.8] | Tamil 54% | 71.8% | 78.6% | – |
| TF-IDF + LR, 4 languages withheld | 86.5% [82.5–90.4] | Tamil 61% | 74.8% | 87.9% | – |
| TF-IDF + LR, all rows | 92.1% [89.2–95.1] | Spanish 79% | 93.2% | 92.7% | – |
| gemma4-31b as classifier | 97.2% [95.3–98.9] | Spanish 86% | 97.1% | 100.0% | – |

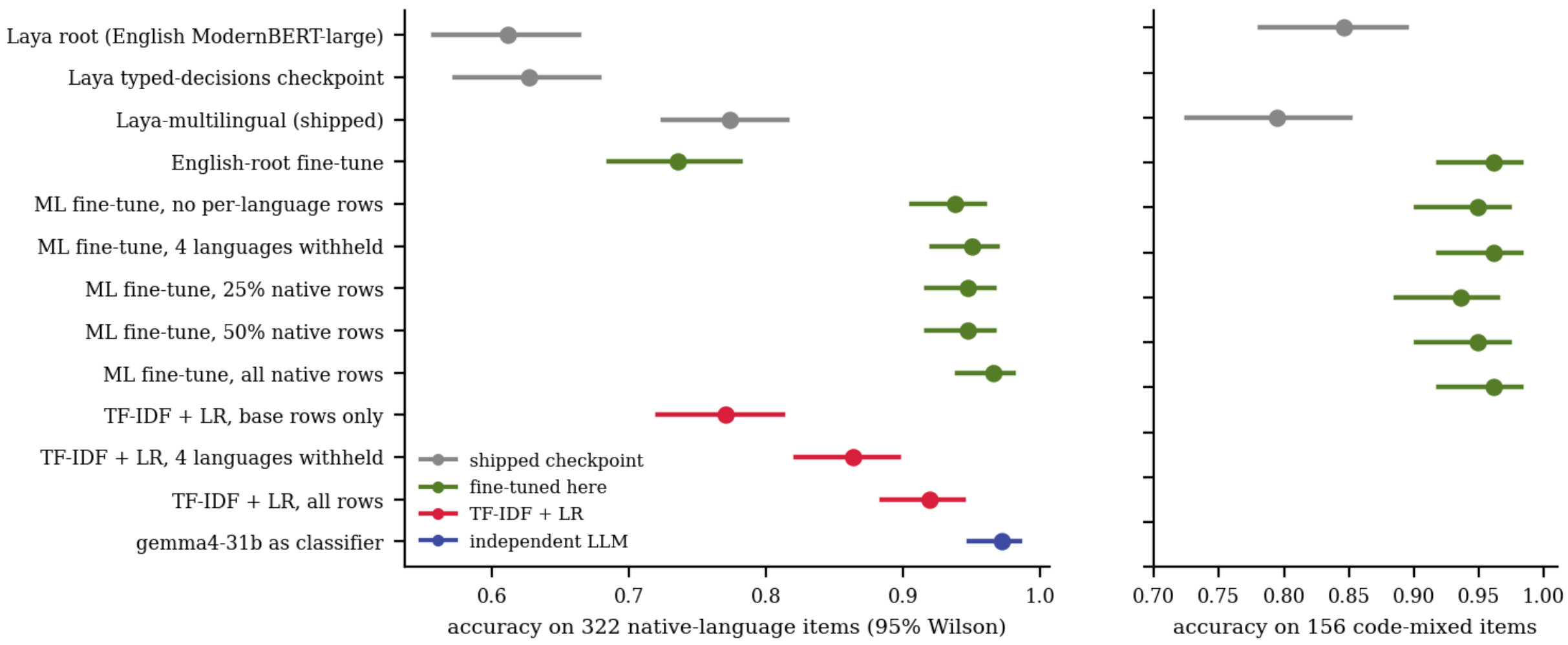

**Figure 8.** The leaderboard as a forest plot: accuracy with 95% intervals on the native split (left) and on the code-mixed split (right; the classical and LLM references were not run on it). The legend groups the systems by family.

## 8.3 Rank stability and pairwise structure

Resampling the test items 2,000 times (Figure 9) shows how well each position is determined. By mean bootstrap rank the order at the top is gemma4-31b (1.4), ML FT 100% (1.7), ML FT w/o 4 langs (3.9), ML FT 50% (4.3), ML FT 25% (4.3); the five fine-tuned multilingual systems and the independent LLM occupy overlapping ranks, whereas the two lowest by mean rank are root and typed-dec.. The pairwise exact tests show the same structure: 55 of 78 pairs are separated at $p<0.01$, and the largest accuracy difference between two systems that are *not* separated is 4.7 points.

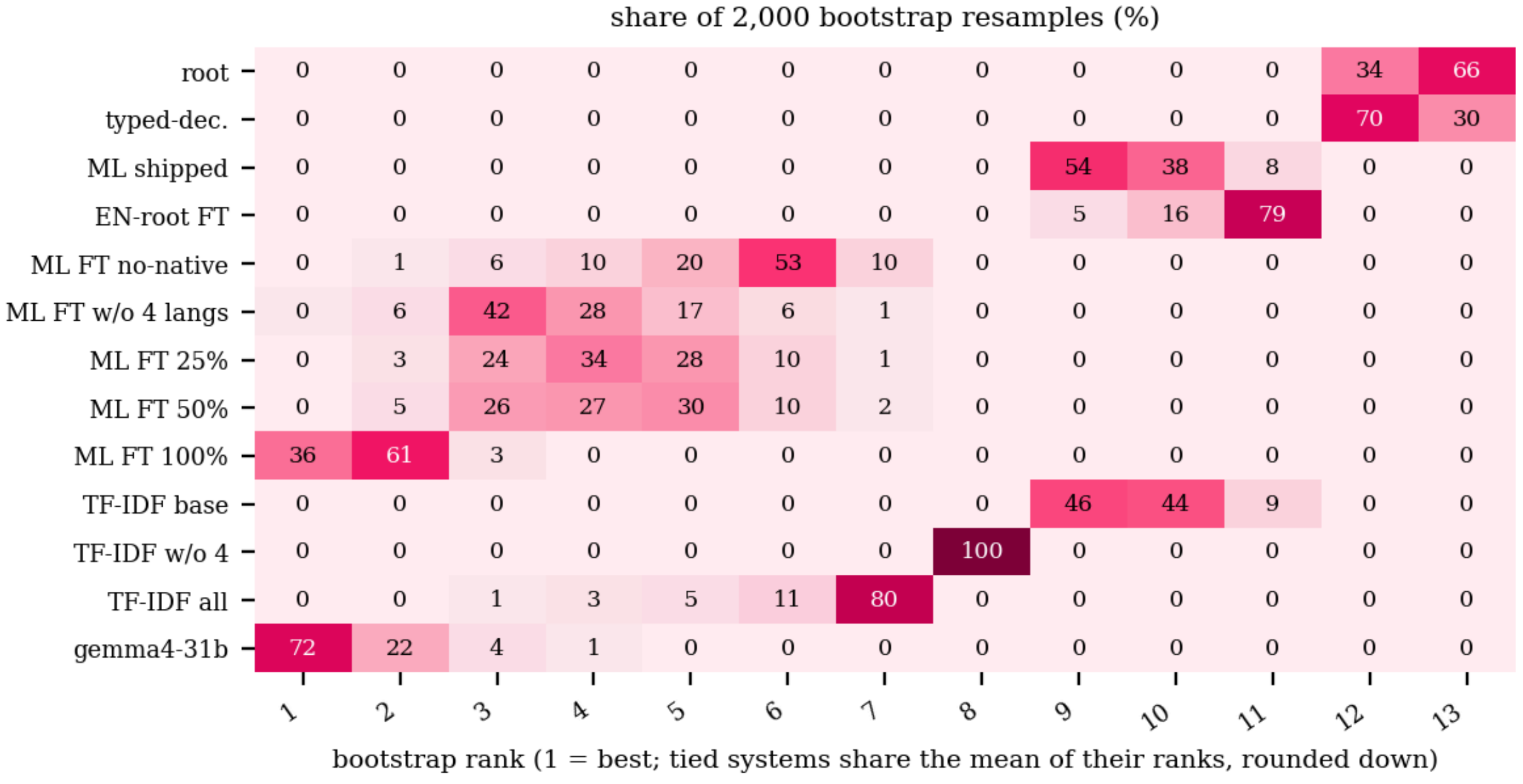

**Figure 9.** Rank stability of the leaderboard under resampling of the test items. Rows are systems, columns are ranks; a sharp row means the position is well determined, a smear means neighbouring systems cannot be ordered on this test size.

## 8.4 Seed variation

Three seeds of the best recipe score 96.6%, 96.9%, 95.7% (standard deviation 0.6 points, range 1.2). Any two seeds answer 2.7% of the items differently, a floor on the discordance that a one-seed comparison cannot get below. For comparison, the 25%, 50% and 100% native-row models differ from one another by 1.9 points; that exceeds the seed-to-seed range, but it is small against the sampling interval, so we treat them as one cluster in the rest of the paper.

## 9 Analyses

### 9.1 Per language

Figure 10 shows every system on every language. The English root ranges from 32% on Arabic to 83% on Japanese; the shipped multilingual checkpoint from 56% (Hindi) to 95% (Portuguese); the best fine-tune from 88% (Bengali) to 100%, with 5 languages at 100%. The largest per-language gain from the root to the shipped multilingual checkpoint is on Arabic (+57 points); the shipped checkpoint is *below* the root on Spanish, Chinese. With about 27 items per cell, one item is 3.7 points and a cell interval is roughly ±11–19 points, so only differences of tens of points should be read as language effects. The weakest writing-system group (Table 12) is Arabic for the root, Indic for the shipped multilingual checkpoint and Indic for the best fine-tune, and fine-tuning narrows the spread between languages.

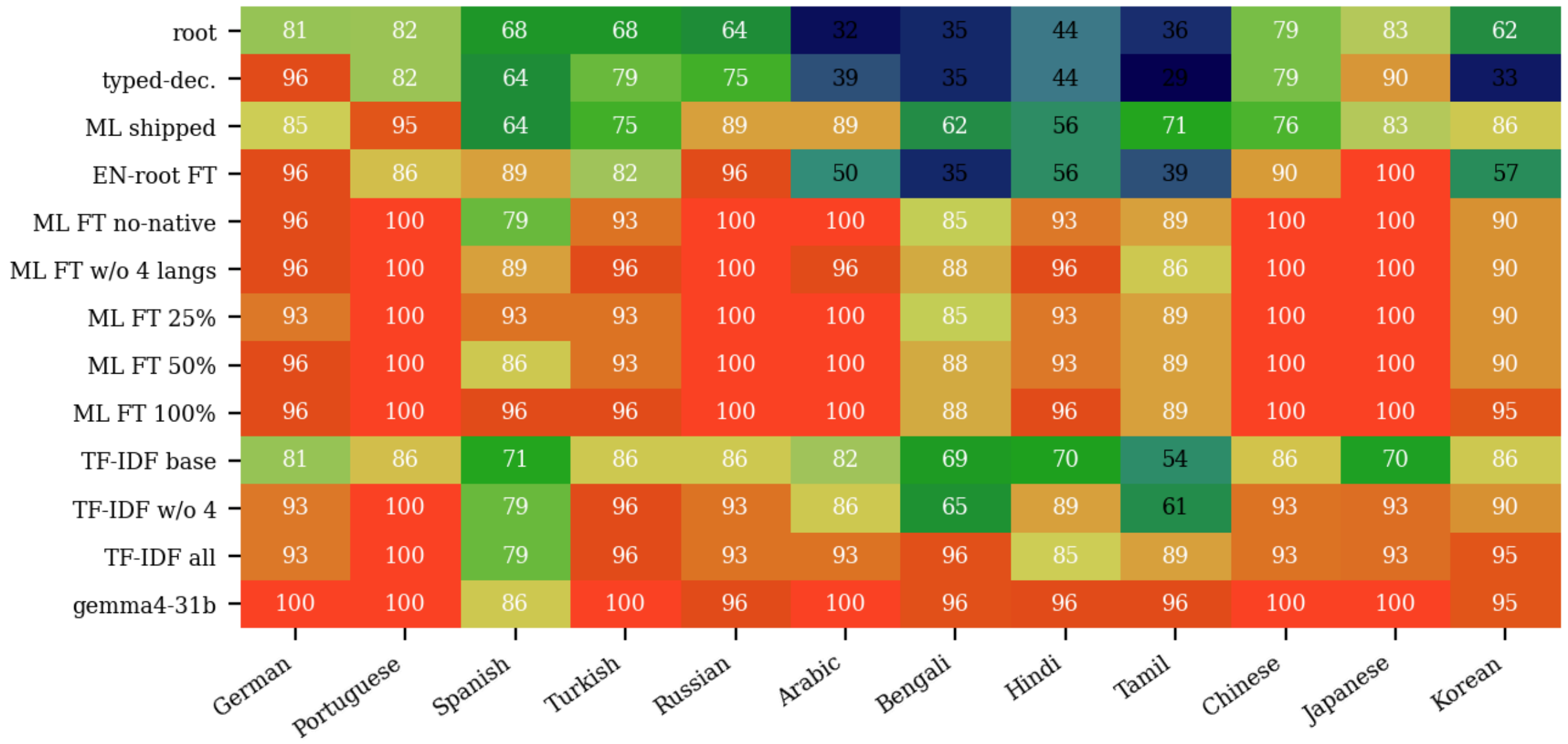

**Figure 10.** Accuracy (%) of every system on every language (about 27 items per cell; one item is worth about 3.7 points, so read differences below ten points as noise).

### 9.2 Per writing system

Pooling languages by writing system (Table 12) reduces the noise: Latin (n=105), Cyrillic (n=28), Arabic (n=28), Indic (n=81), CJK + Hangul (n=80). The shipped checkpoint is weakest on the Indic group (63%), and after fine-tuning the weakest group is Indic (91%). Fine-tuning without any per-language rows already lifts every group to at least 89%.

**Table 12.** Accuracy by writing-system group (pooled over the languages in the group).

| Group | Languages | n | Root | ML shipped | ML FT no-native | ML FT 50% | ML FT 100% |
|---|---|---|---|---|---|---|---|
| Latin | German, Portuguese, Spanish, Turkish | 105 | 74% | 79% | 91% | 93% | 97% |
| Cyrillic | Russian | 28 | 64% | 89% | 100% | 100% | 100% |
| Arabic | Arabic | 28 | 32% | 89% | 100% | 100% | 100% |
| Indic | Bengali, Hindi, Tamil | 81 | 38% | 63% | 89% | 90% | 91% |
| CJK + Hangul | Chinese, Japanese, Korean | 80 | 76% | 81% | 98% | 98% | 99% |

### 9.3 Tokenizer fertility and resource tier

Tokenizer fragmentation is the usual explanation for weak performance on non-Latin scripts [Rust et al. 2021; Ahia et al. 2023; Petrov et al. 2023]. We test it on the benchmark with the standard fertility measure (tokens per character). Across the twelve languages the Spearman correlation between fertility and accuracy is ρ=-0.34 for the root checkpoint with its own tokenizer (permutation p=0.28) and ρ=-0.17 for the shipped multilingual one (p=0.60): moderate negative and negligible negative respectively, and neither is distinguishable from zero at n=12. The ratio of root to multilingual token counts per language correlates with the accuracy gain from root to multilingual at ρ=+0.47. Figure 11 shows why correlations are weak: the three highest-fertility languages under the root tokenizer are Korean (62%), Chinese (79%), Tamil (36%) and the three lowest are Spanish (68%), Portuguese (82%), German (81%), so accuracy does not follow fragmentation monotonically. A moderate negative correlation is compatible with these data and cannot be excluded at

n=12, but fertility alone is not a usable predictor of per-language accuracy on this task; it is also compatible with accounts in which the amount of pre-training data per language matters [Wu and Dredze 2020].

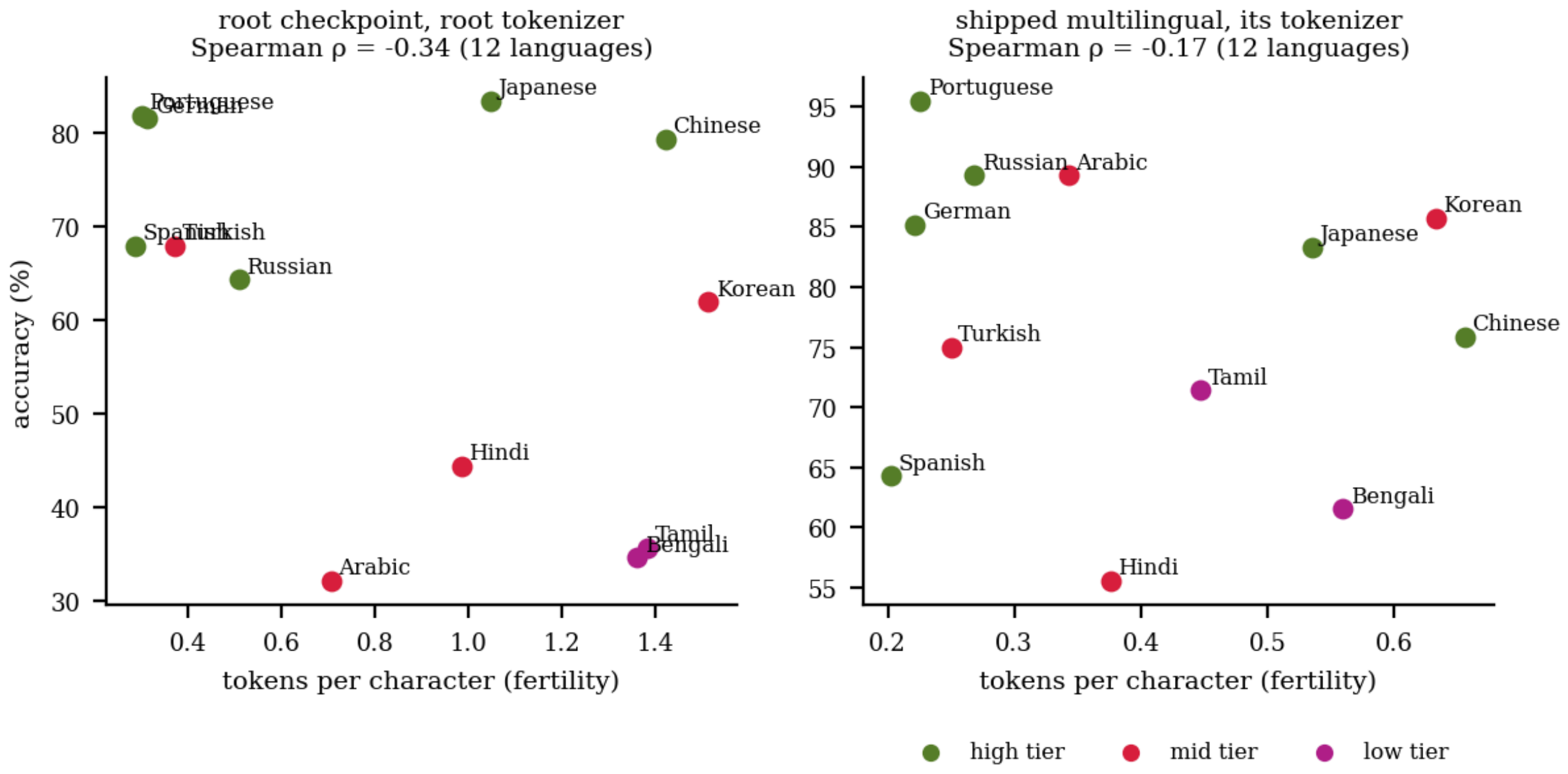

**Figure 11.** Accuracy against tokenizer fertility for each language, coloured by the coarse resource tier used in this paper (an author-assigned grouping, not a measurement).

Grouping the languages by a coarse resource tier (high: German, Portuguese, Spanish, Russian, Chinese, Japanese; mid: Turkish, Arabic, Hindi, Korean; low: Bengali, Tamil; an author-assigned grouping in the spirit of [Joshi et al. 2020], not a measurement) gives Table 13. For the shipped checkpoint the tiers read 82%, 76% and 67% from high to low; after fine-tuning 99%, 97% and 89%. The ordering by tier is preserved after fine-tuning, though the spread shrinks.

**Table 13.** Accuracy by coarse resource tier (author-assigned; see text).

| Tier | Languages | n | Root | ML shipped | ML FT 100% | TF-IDF, all rows |
|---|---|---|---|---|---|---|
| high | German, Portuguese, Spanish, Russian, Chinese, Japanese | 164 | 76% | 82% | 99% | 91% |
| mid | Turkish, Arabic, Hindi, Korean | 104 | 51% | 76% | 97% | 92% |
| low | Bengali, Tamil | 54 | 35% | 67% | 89% | 93% |

## 9.4 Withheld-language transfer

The withheld-language variant (Figure 6) fine-tunes on the base rows plus native rows of eight languages. Table 14 and Figure 12 give the result on Tamil, Bengali, Korean and Arabic. The shipped multilingual checkpoint scores 76.7% on them; the model fine-tuned without them scores 90.3% (+13.6 points, p=0.004); the model fine-tuned on all twelve scores 93.2% (a further +2.9 points, p=0.250). Task fine-tuning therefore transfers to languages without dedicated rows. The additional gain from in-language rows is not significant at this sample size. The model with no per-language rows at all scores 91.3% on the same four languages, so the transfer here comes from the base rows and not from the other eight languages' native rows (-1.0 points for adding the eight languages, p=1.000).

**Caveat on the word 'withheld'.** The base rows contain generic multilingual texts (Section 4.3): 31 in Tamil, 33 in Bengali, 32 in Korean, 34 in Arabic. These are about 40% of the 80 dedicated rows each of those languages has in the full set. The result is thus *reduced exposure*, not zero-shot transfer; an exposure-free comparison would need a base set rebuilt without the generic block. The bag-of-n-grams model shows what withholding does to a model with no pre-training: trained without the four languages it scores 74.8% on them against 93.2% with them.

**Table 14.** Withheld-language transfer: accuracy on the four withheld languages and on the remaining eight, for every system that has a counterpart.

| Languages | n | ML shipped | ML FT no-native | ML FT w/o 4 langs | ML FT 100% | TF-IDF base | TF-IDF w/o 4 | TF-IDF all |
|---|---|---|---|---|---|---|---|---|
| withheld: Tamil, Bengali, Korean, Arabic | 103 | 76.7% | 91.3% | 90.3% | 93.2% | 71.8% | 74.8% | 93.2% |
| remaining eight | 219 | 77.6% | 95.0% | 97.3% | 98.2% | 79.5% | 91.8% | 91.3% |

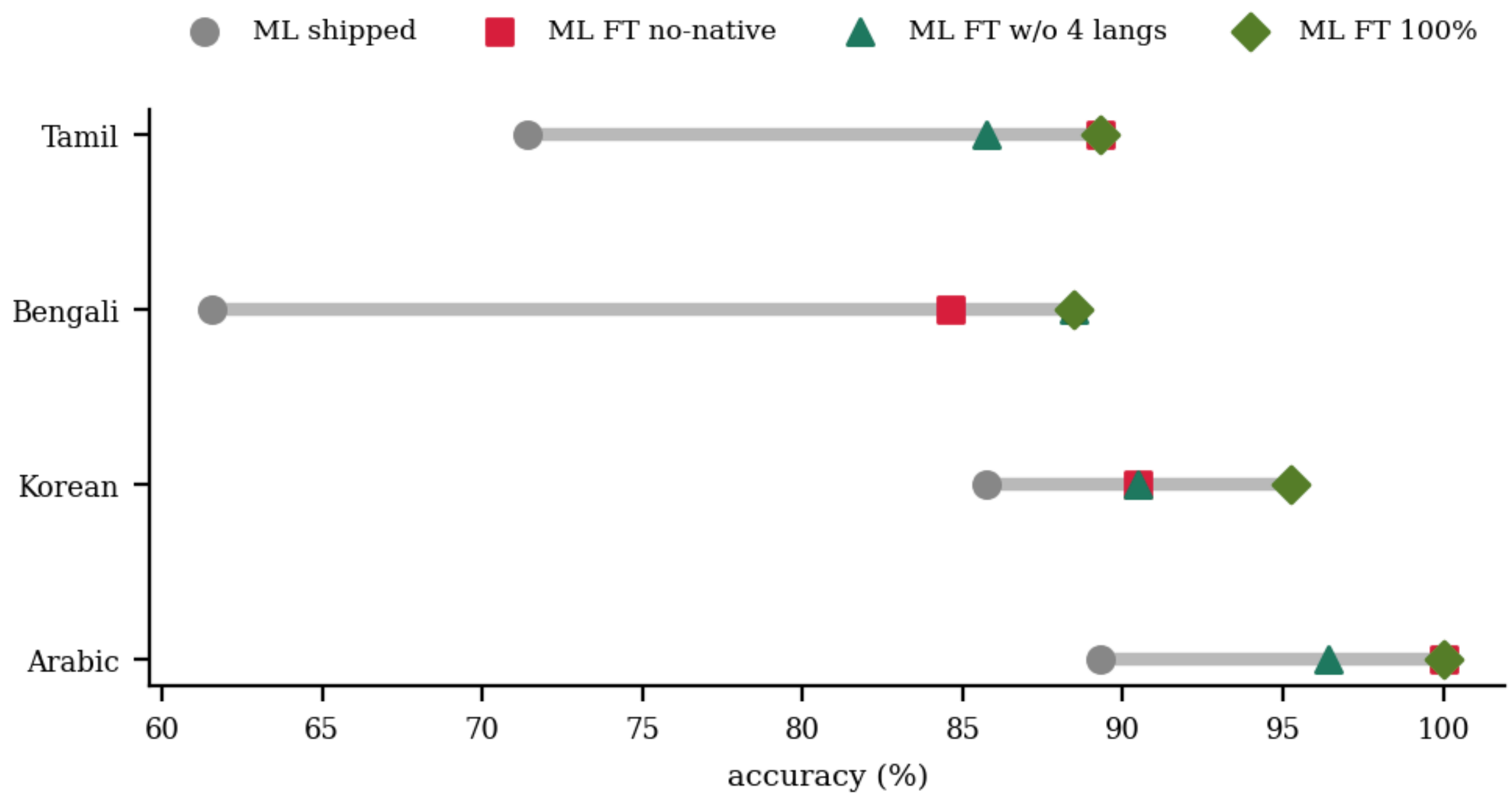

**Figure 12.** The four withheld languages, one row per language: the shipped checkpoint, the model with no per-language rows, the model whose fine-tuning rows exclude these languages, and the model trained on all twelve.

### 9.5 Native-row learning curve

Adding native rows to the base set moves accuracy from 93.8% (no per-language rows) to 94.7% (241 rows), 94.7% (483) and 96.6% (967); Table 15 shows each step against the previous one. Most of the available gain over the shipped checkpoint is already present with *no* dedicated native rows (+16.5 points, $p<0.001$), and the remaining steps are small against the seed-to-seed spread of 0.6 points. A classical n-gram model needs far more rows to get that far: its accuracy first reaches that of the shipped multilingual checkpoint (77.3%) at about 227 native rows (of the 967 available) and it stays below the fine-tuned encoder with all rows (+4.7 points ($p=0.014$)). For a user with a new language the practical reading is that task fine-tuning of the multilingual encoder, not native data volume, is the main lever, and a few hundred rows suffice for the rest.

**Table 15.** Learning curve over native rows (the base rows are constant). The last two columns compare each model with the previous row.

| Native rows | Model | Accuracy [95% CI] | Δ points | McNemar |
|---:|---|---:|---:|---:|
| 0 | ML fine-tune, no per-language rows | 93.8% [90.6–95.9] | – | – |
| 241 | ML fine-tune, 25% native rows | 94.7% [91.7–96.7] | +0.9 | p=0.607 |
| 483 | ML fine-tune, 50% native rows | 94.7% [91.7–96.7] | +0.0 | p=1.000 |
| 967 | ML fine-tune, all native rows | 96.6% [94.0–98.1] | +1.9 | p=0.109 |

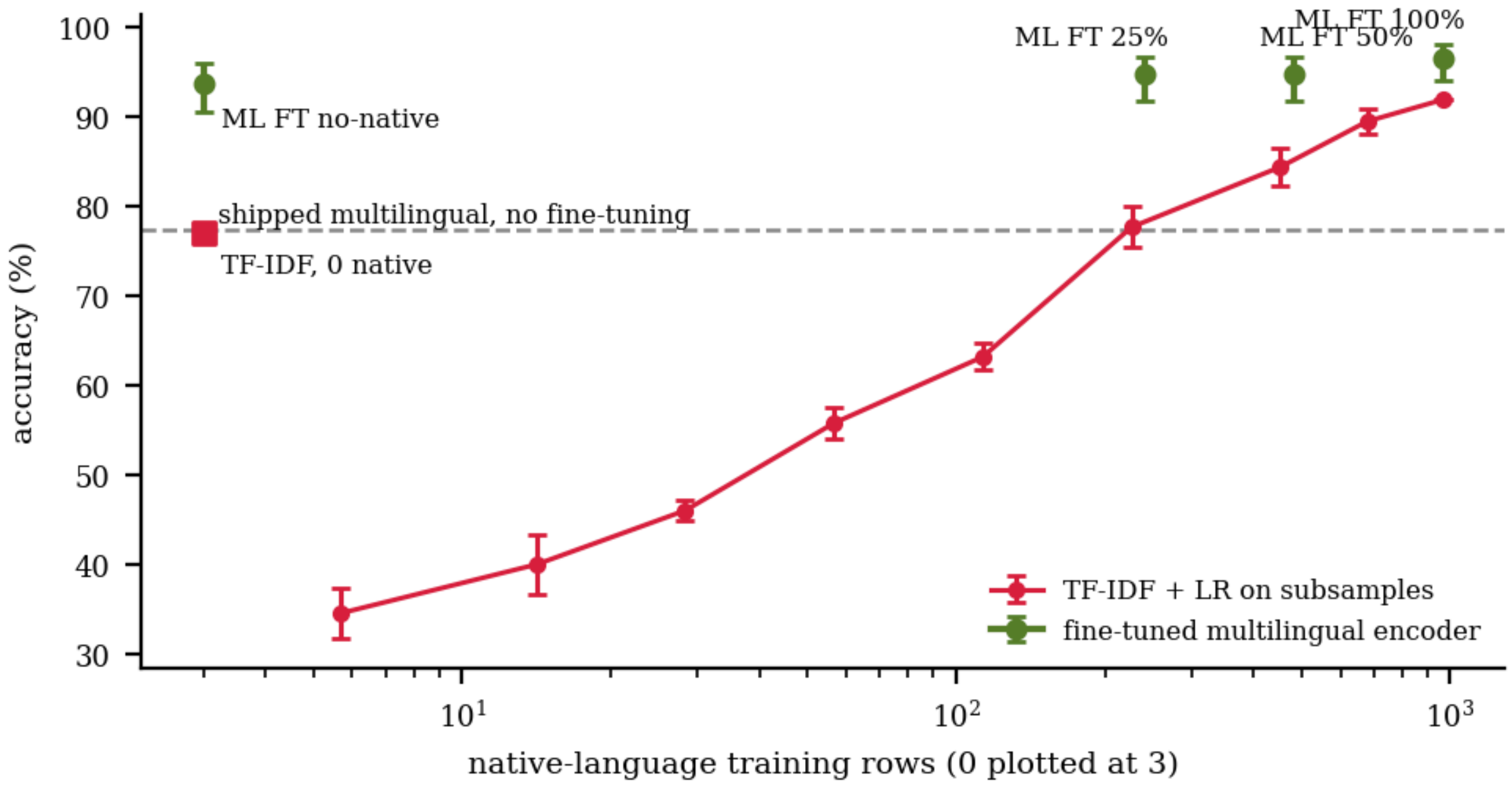

**Figure 13.** Native-row learning curve (legend identifies the families; bars on the fine-tuned points are 95% intervals, bars on the n-gram curve are standard deviations over subsamples). The n-gram classifier's x-position uses the expected native share (967 of 1,707 rows), since it also trains on the base rows.

## 9.6 Code-mixed and romanised text

On romanised and code-mixed text (Table 16, Figure 14) the English root scores 84.6% and the shipped multilingual checkpoint 79.5% (-5.1 points for the multilingual one, p=0.134), whereas on native-script text the multilingual checkpoint is +16.1 points better. The sign of the difference reverses relative to the native split, but the difference is not significant at p<0.01, so we report it as a direction and not as a finding. The shipped multilingual checkpoint is weakest on Arabizi (73%), the variety in which a non-Latin language is written with Latin letters and digits. Fine-tuning closes the gap: the best fine-tune scores 96.2% (+16.7 points over the shipped multilingual checkpoint, p<0.001). Notably the English-root fine-tune also scores 96.2% here although it reaches only 73.6% on the native split: **the code-mixed split measures something the native split does not**, so the two must be reported separately and a high code-mixed score must not be read as multilingual competence. At 156 items the 95% intervals are about ±3–8 points, so we only draw conclusions about differences larger than that.

**Table 16.** Code-mixed accuracy by variety, with the native-split accuracy of the same systems for reference. Intervals for the pooled row are 95% Wilson.

| Variety (n) | root | ML shipped | ML FT no-native | ML FT w/o 4 langs | ML FT 100% | EN-root FT |
|---|---|---|---|---|---|---|
| Hinglish (58) | 74% | 79% | 93% | 95% | 93% | 90% |
| Spanglish (47) | 94% | 87% | 94% | 98% | 98% | 100% |
| Arabizi (51) | 88% | 73% | 98% | 96% | 98% | 100% |
| all code-mixed (156) | 84.6% [78.1–89.4] | 79.5% [72.5–85.1] | 94.9% [90.2–97.4] | 96.2% [91.9–98.2] | 96.2% [91.9–98.2] | 96.2% [91.9–98.2] |
| native split (322) | 61.2% | 77.3% | 93.8% | 95.0% | 96.6% | 73.6% |

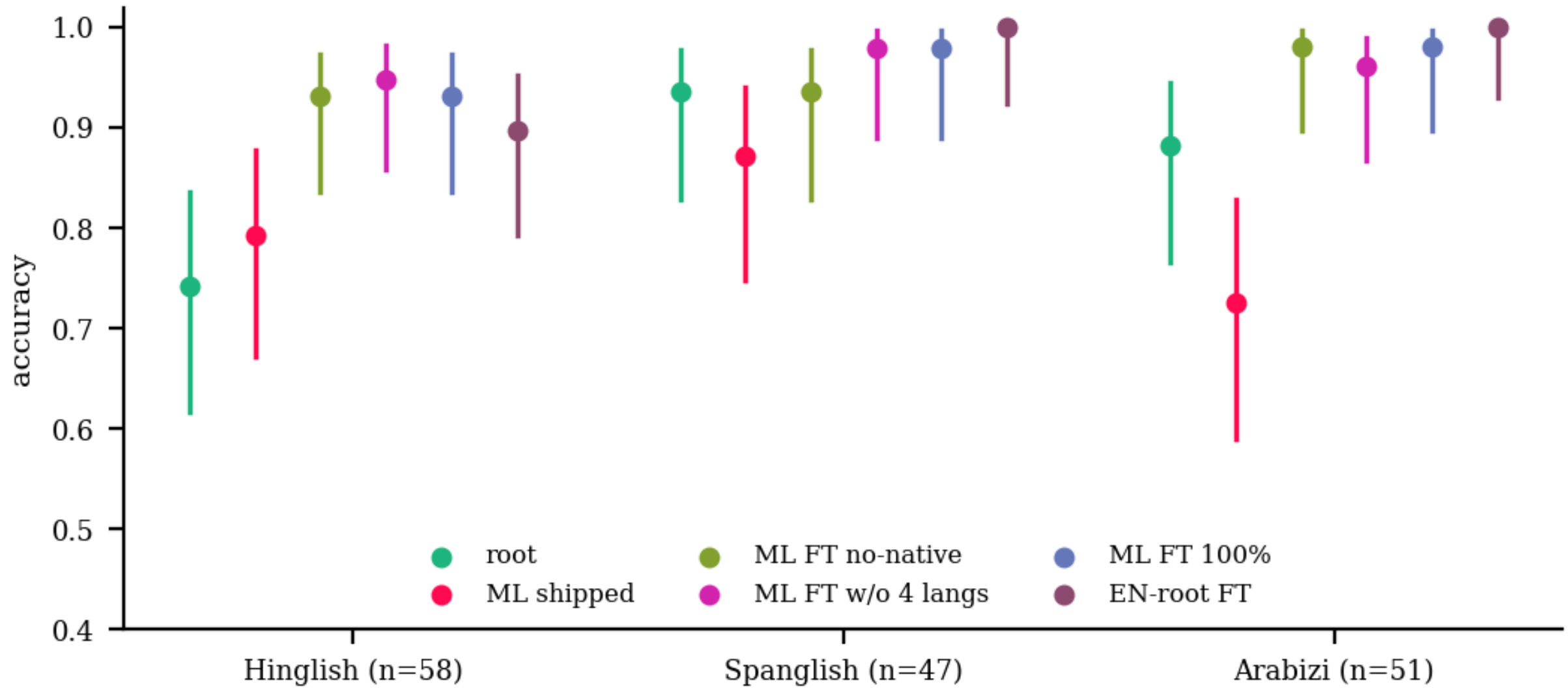

**Figure 14.** The code-mixed split by variety: accuracy with 95% intervals for each system (points are offset horizontally within a variety).

## 9.7 The shipped Router

The family's Router detects script and language and sends non-English text to the multilingual checkpoint. On the native split it sends 313 of 322 items there and 9 to the English root; routed accuracy is 77.3% [72.4–81.6], against 77.3% for the shipped multilingual checkpoint used directly (Table 17, Figure 15). The misrouted items are 9 Turkish; all are in a language written in Latin letters, and we did not investigate the Router's detection rule. On those 9 items the root scores 100% and the multilingual checkpoint 100%, so the misrouting costs nothing on this sample. All 75 English items stay on the English root (77.3% [66.7–85.3]). Because a deployment through the Router never exposes the root to native-script text, the root-versus-multilingual gap (Table 10) overstates what a Router user would see: the relevant comparison for such a user is the routed system against a fine-tuned multilingual model (+35.4 points for the latter over the root; +19.3 points over the routed system).

**Table 17.** The shipped Router per language: share of items sent to the multilingual checkpoint, accuracy of the routed system, and accuracy of each checkpoint used directly. The last row has 75 English department items.

| Language | n | Routed to multilingual | Router accuracy | ML direct | Root direct |
|---|---|---|---|---|---|
| German | 27 | 100% | 85% | 85% | 81% |
| Portuguese | 22 | 100% | 95% | 95% | 82% |
| Spanish | 28 | 100% | 64% | 64% | 68% |
| Turkish | 28 | 68% | 75% | 75% | 68% |
| Russian | 28 | 100% | 89% | 89% | 64% |
| Arabic | 28 | 100% | 89% | 89% | 32% |
| Bengali | 26 | 100% | 62% | 62% | 35% |
| Hindi | 27 | 100% | 56% | 56% | 44% |
| Tamil | 28 | 100% | 71% | 71% | 36% |
| Chinese | 29 | 100% | 76% | 76% | 79% |
| Japanese | 30 | 100% | 83% | 83% | 83% |
| Korean | 21 | 100% | 86% | 86% | 62% |
| English items | 75 | none (all to English root) | 77.3% [66.7–85.3] | – | – |

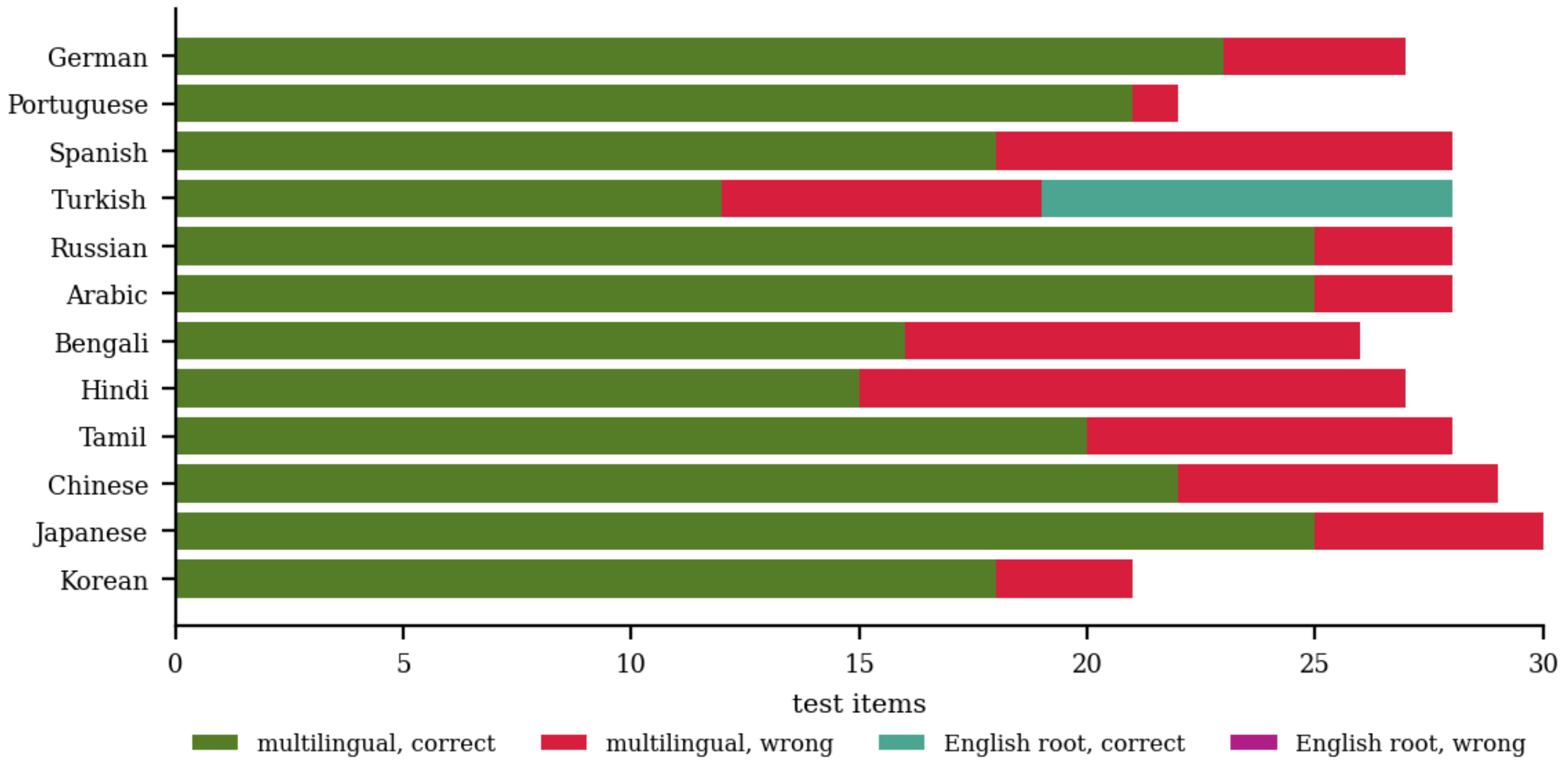

**Figure 15.** What the shipped Router does with each language: which checkpoint receives the item and whether the answer is correct.

# 10 Stress conditions and confidence

## 10.1 Stress conditions

Two protocol choices could change a score without changing the model: the order in which the options are listed and the way they are named. Table 18 and Figure 16 quantify both on the native split. Cycling the option order changes the answer for 37% of root predictions, 8% for the shipped multilingual checkpoint and 1.6% for the best fine-tune, so a fine-tuned system is much less order-sensitive, but only a system with a flip rate near zero would make the canonical order irrelevant. Replacing the names by letters, numbers or opaque tokens lowers accuracy of the best fine-tune from 96.6% to 91.6–94.4%; rotating the names while keeping the descriptions drops it to 6.5%, and the model follows the *name* of the gold option on 92% of items. The fine-tuned systems read the option names, not only the descriptions, and the benchmark question must therefore be used exactly as published (rule 1 of Section 6.3) [Pezeshkpour and Hruschka 2023; Zheng et al. 2023].

**Table 18.** Stress conditions on the native split (accuracy in %, same 322 items). ‘Rotated’ keeps the three descriptions in place and cyclically shifts the names; the last column is the share of rotated items on which the model follows the *name* of the gold option instead of its description.

| System | Original | Letters | Numbers | Opaque names | Rotated names | Over 3 option orders | Answers that change with order | Follows rotated name |
|---|---|---|---|---|---|---|---|---|
| Laya root (English ModernBERT-large) | 60.9 | 60.6 | 59.9 | 57.8 | 32.0 | 59.7 | 37.3% | 45% |
| Laya-multilingual (shipped) | 77.3 | 75.8 | 72.4 | 68.9 | 13.7 | 76.2 | 7.8% | 70% |
| ML fine-tune, no per-language rows | 93.8 | 87.6 | 90.7 | 81.7 | 8.1 | 93.6 | 3.4% | 86% |
| ML fine-tune, all native rows | 96.6 | 93.5 | 94.4 | 91.6 | 6.5 | 96.3 | 1.6% | 92% |

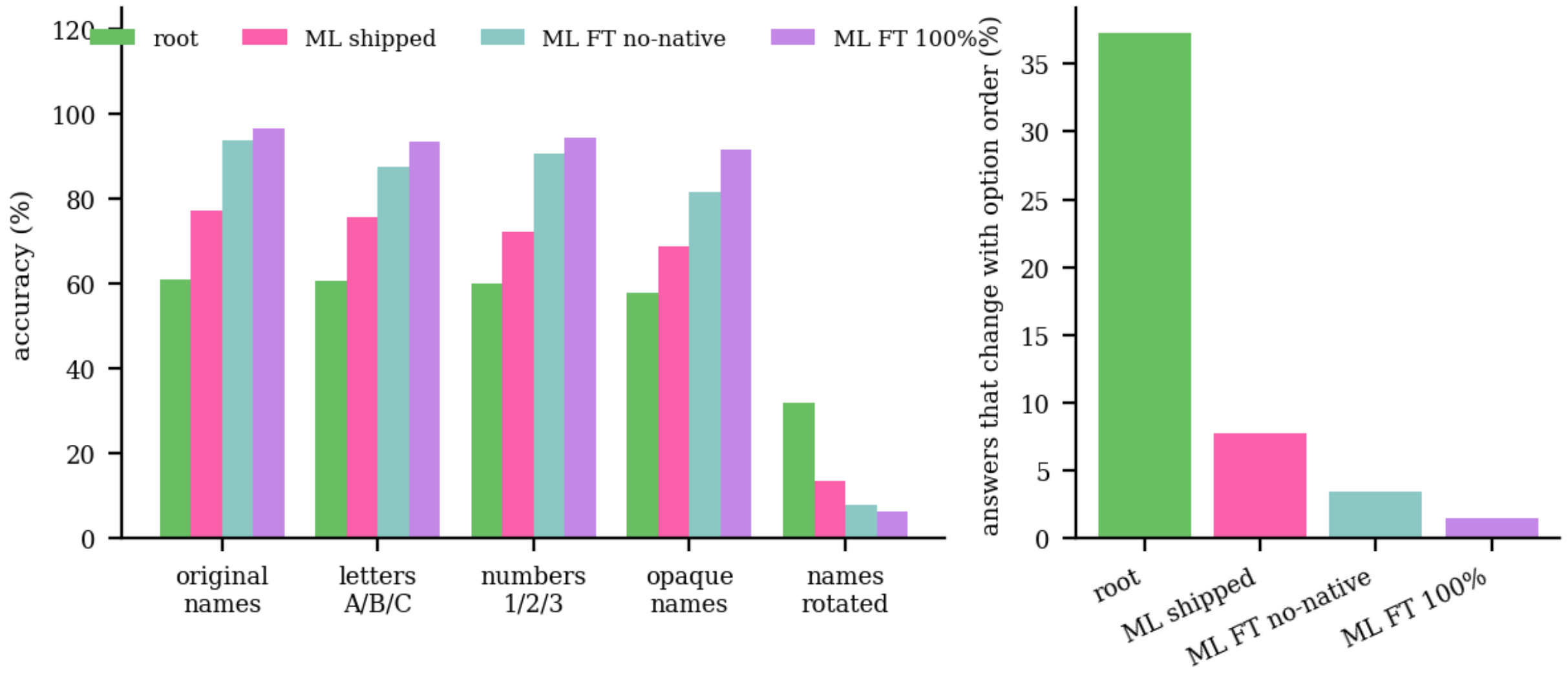

**Figure 16.** Stress conditions on the native split. Left: accuracy when the option names are replaced by letters, numbers or opaque tokens, or rotated while the descriptions stay in place. Right: share of items whose predicted label changes when the three options are cyclically re-ordered.

## 10.2 Confidence and abstention

A router is only useful if it knows when it is wrong [Guo et al. 2017; Geifman and El-Yaniv 2017]. The root checkpoint's mean confidence is 59.5% against an accuracy of 61.2%; the shipped multilingual checkpoint is 84.7% confident at 77.3% accuracy; the best fine-tune 97.7% at 96.6% (Table 19, Figure 17). Selective accuracy rises when low-confidence items are withheld: the shipped multilingual checkpoint goes from 77.3% to 83.3% on its most confident 80%, the root from 61.2% to 67.3%. Fine-tuning also makes the confidence more informative about errors (lower AURC and ECE). A caveat for users of the benchmark: with 11 errors, the best system's calibration is estimated from very few mistakes, and ECE in particular is noisy at that count.

**Table 19.** Confidence quality on the native split. AURC is the area under the risk-coverage curve (lower is better); 'acc@80%' is accuracy on the 80% most confident items; the last column counts errors made with confidence of at least 0.9.

| System | Accuracy | Mean confidence | ECE | NLL | AURC | acc@80% | Confident errors |
|---|---|---|---|---|---|---|---|
| Laya root (English ModernBERT-large) | 61.2% | 59.5% | 0.058 | 0.806 | 0.171 | 67.3% | 0 of 125 |
| Laya-multilingual (shipped) | 77.3% | 84.7% | 0.078 | 0.653 | 0.097 | 83.3% | 17 of 73 |
| ML fine-tune, no per-language rows | 93.8% | 94.1% | 0.038 | 0.204 | 0.012 | 98.8% | 5 of 20 |
| ML fine-tune, all native rows | 96.6% | 97.7% | 0.011 | 0.125 | 0.010 | 98.8% | 6 of 11 |

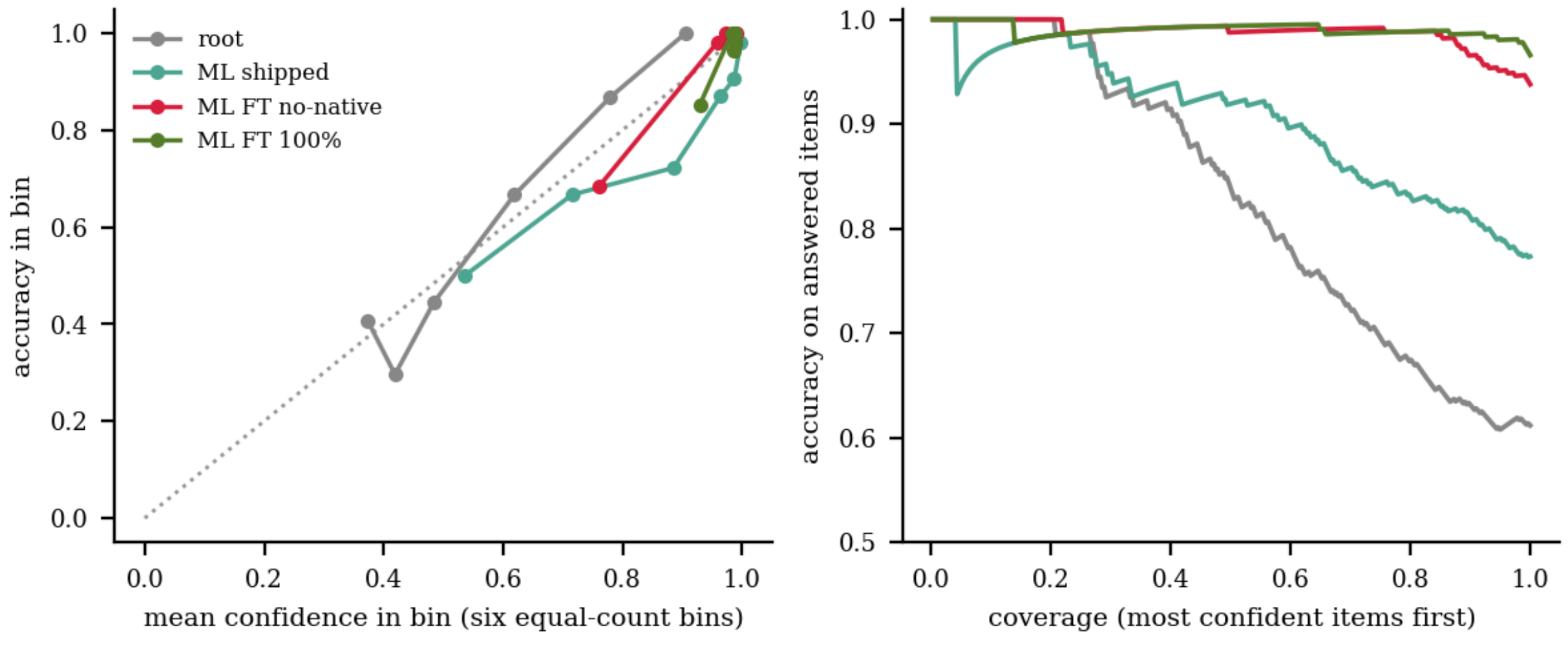

**Figure 17.** Confidence quality on the native split. Left: reliability curves (a curve below the diagonal is over-confident). Right: accuracy as the model abstains on its least confident items.

## 11 Error taxonomy and item difficulty

Errors fall in three classes (Table 20, Figure 18, right panel): items whose gold label the independent labeller disputes; confusions between the two substantive departments; and confusions that involve the residual ‘other’ department. For the shipped multilingual checkpoint 54 of 73 errors involve ‘other’ and 11 are billing/technical confusions; for the best fine-tune 6 of 11 involve ‘other’, 3 are billing/technical, and 2 sit on disputed gold labels. The residual class is the hardest to separate for both the models and the labellers: it is defined by exclusion, so a message about a tangential billing or technical issue is genuinely borderline.

**Table 20.** Error taxonomy: every error of every Laya-based system is assigned to one class.

| System | Errors | Disputed gold | Billing / technical confusion | Involves ‘other’ | Confident (≥0.9) |
|---|---|---|---|---|---|
| Laya root (English ModernBERT-large) | 125 | 8 | 37 | 80 | 0 |
| Laya typed-decisions checkpoint | 120 | 6 | 9 | 105 | 0 |
| Laya-multilingual (shipped) | 73 | 8 | 11 | 54 | 17 |
| English-root fine-tune | 85 | 3 | 35 | 47 | 36 |
| ML fine-tune, no per-language rows | 20 | 8 | 2 | 10 | 5 |
| ML fine-tune, all native rows | 11 | 2 | 3 | 6 | 6 |

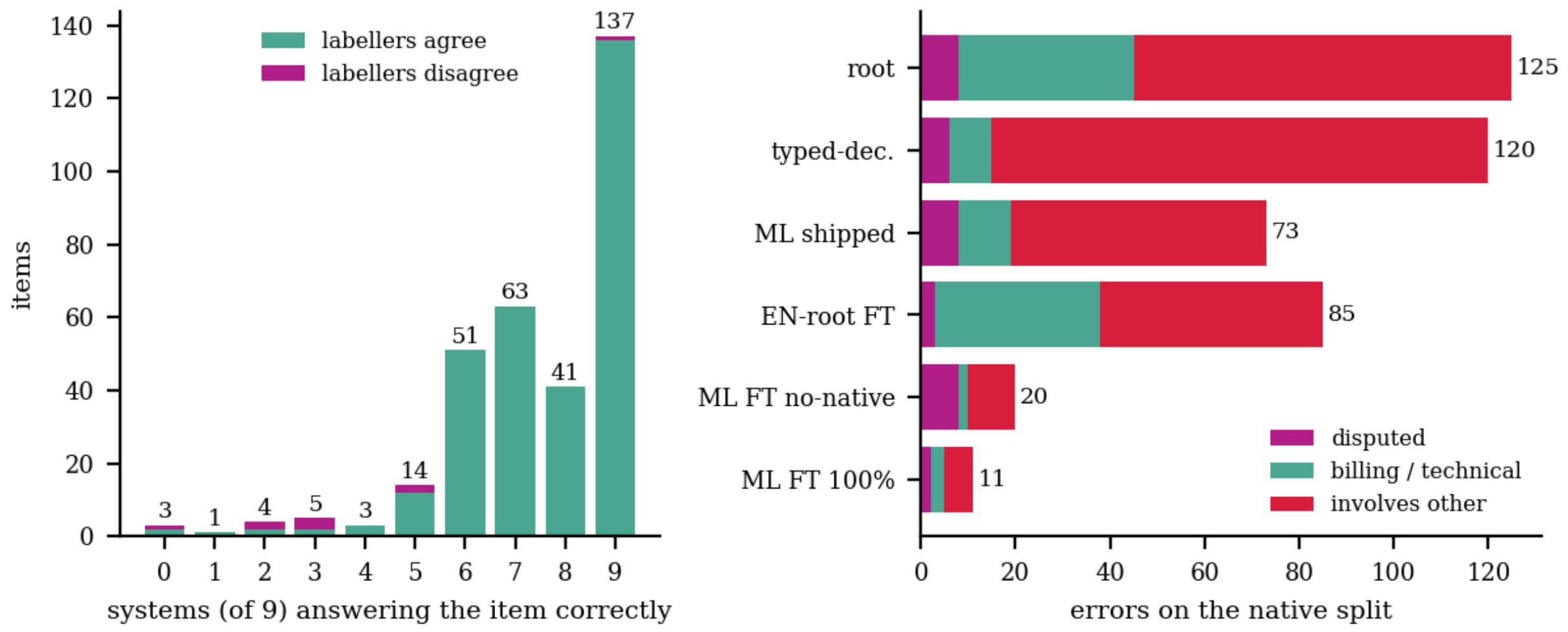

**Figure 18.** Item difficulty and error taxonomy. Left: how many of the 9 Laya-based systems solve each item (right-most bar: solved by all; left-most: by none; the upper segment counts items whose gold label the independent labeller disputes). Right: errors of each system by class — disputed gold label, confusion between the two substantive departments, or confusion involving the residual ‘other’ department.

### 11.1 Item difficulty

Of the 322 items, 137 (43%) are answered correctly by all nine Laya-based systems and 3 by none; 4 are solved by at most one (Figure 18, left panel). Of the 3 items that no system solves, 1 is disputed by the independent labeller, so the unsolved items are not mainly items whose gold label is in doubt. Table 21 lists the hardest items; the benchmark user should read them before concluding that a model is at fault.

**Table 21.** The hardest items: answered correctly by at most one of the nine Laya-based systems.

| Language | Message | Gold | Indep. labeller | Systems correct |
|---|---|---|---|---|
| Bengali | আমি গত সপ্তাহে ল্যাপ্টপটি কেনা করেছিলাম কিন্তু এখনও পর্যন্ত কোনো ইনভয়েস ইমেইলে পাইনি, দয়া করে তা পাঠান। | billing | billing | 0 |
| Tamil | பிளாட் டீக் கடந்த ஐந்து நிமிடங்களாக ஓப் Abe donnant பொறி இயங்கவில்லை. | technical | other | 0 |
| Tamil | டிராகன் பல்லியில் ஸ்க்ரீன் கருமை நிறமாகி எதையும் காட்டாது. | technical | technical | 0 |
| Tamil | கோவேப் சேவை இன்று அதிகாலையிலிருந்து சரியாக இயங்கவில்லை. | technical | technical | 1 |

## 11.2 Residual errors of the best system

**Table 22.** Every error of the best fine-tuned system on the native split (texts shortened; […] marks an elided word). Columns: independent labeller, shipped multilingual checkpoint, best fine-tune with its confidence, and the error class.

| Language | Message | Gold | Indep. | ML shipped | ML FT (conf.) | Class |
|---|---|---|---|---|---|---|
| German | Der Mieter im Erdgeschoss hat trotz mehrfacher Mahnungen die Miete nicht gezahlt. Welche rechtlichen Schritte sind noch… | other | other | billing | billing (0.58) | other-boundary |
| Bengali | আমার পছন্দের গানটি বাজানোর সময় গিটারের স্ট্রিংয়ে একটি অস্বাভাবিক ধ্বনি আসছে, যা ঠিক না হলে ইনস্ট্রুমেন্টটির মূল্য হ্রা… | other | other | other | technical (0.96) | other-boundary |
| Bengali | আমি গত সপ্তাহে ল্যাপ্টপটি কেনা করেছিলাম কিন্তু এখনও পর্যন্ত কোনো ইনভয়েস ইমেইলে পাইনি, দয়া করে তা পাঠান। | billing | billing | technical | other (0.98) | other-boundary |
| Tamil | பிளாட் டீக் கடந்த ஐந்து நிமிடங்களாக ஓப் Abe donnant பொறி இயங்கவில்லை. | technical | other | other | other (0.75) | disputed gold |
| Korean | 결제 시스템에서 카드 검증 단계에서 예외가 던져져 주문이 완료되지 못하고 있습니다. | technical | technical | technical | billing (0.98) | billing/technical |
| Bengali | আমার অর্ডারের জন্য পেমেন্ট নিশ্চিত করা হয়েছে কিনা? ব্যাংক স্টatement-এ ট্রানজেকশন দেখা যাচ্ছে কিন্তু অ্যাপে স্ট্যাটাস আ… | billing | billing | technical | technical (0.58) | billing/technical |
| Turkish | Security patch failed to install, rollback successful but vulnerability remains, need hotfix | technical | technical | technical | other (0.98) | other-boundary |
| Tamil | டிராகன் பல்லியில் ஸ்க்ரீன் கருமை நிறமாகி எதையும் காட்டாது. | technical | technical | other | other (0.99) | other-boundary |
| Spanish | El API de pagos devuelve siempre un error 500 cuando se intenta procesar una transacción superior a cien euros. | technical | technical | billing | billing (0.81) | billing/technical |
| Tamil | கோவேப் சேவை இன்று அதிகாலையிலிருந்து சரியாக இயங்கவில்லை. | technical | technical | technical | other (0.53) | other-boundary |
| Hindi | कंपनी के वित्तीय खातों में गड़बड़ी का शक है, उसे तुरंत जांचा जाना चाहिए। | other | billing | billing | billing (0.98) | disputed gold |

The 11 residual errors of the best fine-tune are concentrated in Bengali (3), Tamil (3), German (1). We deliberately keep them in the benchmark rather than cleaning them: removing the items a strong model gets wrong would inflate every score.

## 11.3 Where the shipped checkpoint is confidently wrong

The shipped multilingual checkpoint makes 17 confident errors (confidence ≥0.9) among its 73, and the best fine-tune corrects 16 of them. These are the cases a deployment would route wrongly without any warning from the confidence score.

# 12 Recommended usage and pitfalls

- **Report the two splits separately.** The English-root fine-tune and the shipped multilingual checkpoint score 73.6% and 77.3% on the native split (3.7 points apart) but 96.2% and 79.5% on the code-mixed split (16.7 points apart). A single pooled number would hide that the two splits measure different things.
- **Do not rank systems within about 4 points of each other on the native split.** The split cannot resolve it (Section 7); use several seeds (0.6 points of seed standard deviation here) and the paired bootstrap.
- **Do not make per-language claims from this split.** About 26 items per language give cell intervals of roughly ±11–19 points. Pool by script group or tier if a language-level statement is needed.
- **Keep the question text and option names fixed.** Rotating the names while keeping the descriptions reduces the best fine-tune to 6.5% (Table 18); changing the schema changes the task.
- **Read the withheld-language track as reduced exposure.** The base rows hold 399 generic multilingual texts (Section 4.3).
- **Compare with a bag of n-grams.** TF-IDF + LR on all rows reaches 91.9%; a multilingual-encoder result is informative only to the extent that it exceeds what lexical cues give.
- **Use the consensus subset as a sanity check, not as the headline.** It removes 9 disputed items and changes the best system's accuracy from 96.6% to 97.1%.

- **Do not tune on the test items.** Both splits are small enough to be over-fitted by repeated model selection [Bowman and Dahl 2021].

## 13 Limitations, ethics and datasheet

Table 23 condenses the datasheet; the paragraphs below spell out the limitations that bear most on how the benchmark should be used.

**Table 23.** Datasheet (condensed).

| Section | Question | Answer |
|---|---|---|
| Motivation | Why was it created? | To give typed-decision encoders a fixed, documented multilingual routing test, with baselines, instead of ad-hoc spot checks. |
| Motivation | Who created it? | The authors, as part of an evaluation of a public typed-decision model family; no external dataset was used. |
| Composition | What are the instances? | A short support message in one of 12 languages (or code-mixed) with one gold department out of three. Test: 322 native + 156 code-mixed; optional training: 967 native rows. |
| Composition | Is there personal or sensitive data? | No. All texts are generated by a language model from a topical prompt; no real users, names or accounts are involved, although generated texts may contain invented names and amounts. |
| Composition | Are labels reliable? | They are LLM-made and self-consistency filtered; an independent model family agrees on 97.2%. No human annotation was done. |
| Collection | How was it collected? | Teacher-model generation per language and department, blind relabelling by the same model, exact-text de-duplication against other splits (Section 3). |
| Preprocessing | Was anything removed? | Candidates whose blind label disagreed with the intended label (131 of 1420 native-run candidates, 22 of 180 code-mixed candidates). No other cleaning; messages are used verbatim. |
| Uses | Recommended uses | Model comparison on routing-style decisions across languages; transfer and robustness studies; calibration studies. |
| Uses | Uses to avoid | Estimating real-world performance in a language; ranking languages; training production systems without additional human-written data; any claim about a speaker community. |
| Distribution | Where are the files and how to verify them? | In the project's `data` directory as JSON-lines; SHA-256 checksums are in Appendix C. Check the terms of the teacher model before redistributing generated text. |
| Maintenance | Will it be updated? | The files are frozen with the results; a human-written replacement for at least the test split is the intended next version. |

### 13.1 Limitations

- **All labels and texts are LLM-made.** The labels were produced and filtered by one model family; the independent audit (97.2%) bounds but does not remove the dependence, and agreement between two language models is not agreement with speakers [Northcutt et al. 2021; Zheng et al. 2023]. The text itself may be more regular and more standard than what customers write; dialect, typos, scripts mixed within a word and domain vocabulary are under-represented.
- **Machine-written text in low-resource languages.** The teacher's fluency and cultural fit probably vary across the twelve languages; no native speaker has reviewed the texts. This matters most for Tamil and Bengali.
- **One task, one schema.** Routing is topical and has a high ceiling (97.2% for the best entry); the benchmark says little about tasks where syntax or negation decides.
- **Small cells.** About 26 items per language and 156 code-mixed items: only large gaps are resolvable (Section 7).
- **Exposure in the base rows.** 399 generic multilingual rows are shared by all training tracks.
- **Other limits.** Generated text contains artefacts (foreign-script fragments, English text filed under a Latin-script language; Section 4.4); only the best recipe has three seeds (the others are single runs); the resource tiers are an author-assigned grouping, not a measurement; and no classical or LLM baseline was run on the code-mixed split.

### 13.2 Ethical considerations

The data contain no real personal information. The main risk is over-reading: a benchmark of synthetic text could be used to certify a system for lower-resource-language speakers who were never represented in it, and a router that is wrong more often for some languages would then deliver slower or wrong service to those users. We therefore ask that results be reported with the intervals, the split names and the withheld-language caveat above, and that deployments be validated on human-written messages from the target community. Generated text can also carry the teacher's stereotypes about languages and countries; we did not audit for that. Compute cost is small: the benchmark runs in seconds on one GPU (2.0 ms per item for the best encoder).

## 14 Conclusion

PolyRoute turns the question 'does this typed-decision encoder work outside English?' into a measurement with a documented provenance: 322 native-language messages in 12 languages, 156 code-mixed ones, an independent label audit (97.2% agreement), a leakage audit (maximum similarity 0.33), five nested training sets and thirteen leaderboard entries. The first results are that the English root scores 61.2% and the shipped multilingual checkpoint 77.3%; fine-tuning the multilingual encoder reaches 96.6% and most of that gain needs no dedicated native rows; a bag of n-grams is within 5 points of the best fine-tune; and on code-mixed text the English root is nominally ahead of the shipped multilingual checkpoint (85% vs 79%, p=0.134). The split cannot separate systems closer than about 4 points, and its labels, texts and languages come from language models; those limits, not the leaderboard, are the main message for anyone who builds on it.

# Appendix A  Sample items

**Table 24.** The first test item of every language (department cycling through billing, technical, other), shortened to 150 characters.

| Language | Script | Gold | Message |
|---|---|---|---|
| German | Latin | billing | Ich bitte um einen Auszug meiner letzten drei Monatsabrechnungen, um die Details meiner Überweisungen zu verifizieren. |
| Portuguese | Latin | technical | Detectamos uma falha crítica no banco de dados que causou a perda de integridade das tabelas de logs, exigindo uma intervenção imediata do time de inf… |
| Spanish | Latin | other | Pido el mantenimiento preventivo de los equipos de refrigeración de la cámara frigorífica, pues los compresores están emitiendo ruidos anormales y vib… |
| Turkish | Latin | billing | Merhaba, geçen ayki aboneliğim için gönderilen faturanın bir kopyasını e-posta adresime iletebilir misiniz? Ödeme tarihimle ilgili de bilgi almak isti… |
| Russian | Cyrillic | technical | Веб-интерфейс не грузит стили, аудит консоли браузера показывает 404 ошибки при загрузке CSS-файлов с CDN. |
| Arabic | Arabic | other | لديّ سؤال بخصوص جدولة رحلة السفر إلى سويسرا في شهر يوليو، وما هي أفضل الفنادق التي توفر إطلالات جبلية رائعة؟ |
| Bengali | Bengali | billing | ক্যাশ অন ডেলিভারি দিতে গিয়ে ভুল করে বেশি টাকা দিয়েছি, ব্যালেন্স ফেরত পেতে চাই, রসিদসহ যুক্ত করে দিলাম। |
| Hindi | Devanagari | technical | इंटरफ़ेस में बटन क्लिक होने पर कोई रिस्पॉन्स नहीं आ रहा है, जावास्क्रिप्ट कंसोल में 'ReferenceError' दिख रहा है। |
| Tamil | Tamil | other | என் குழந்தையின் பள்ளி அரங்கத்தில் நடந்த கலை விழாவில் எடுக்கப்பட்ட படங்களை எதிர்காலத்தில் பயன்படுத்த சுமதிளதில் சிக்கல்கள் உள்ளன. |
| Chinese | Han | billing | 我的订单状态显示“已发货”，但一直未收到物流信息，且支付记录显示交易成功，请核对账单状态。 |
| Japanese | Han + kana | technical | データベースの同期が完全に停止しており、最新の顧客データが反映されていません。緊急対応を要します。 |
| Korean | Hangul | other | 오버클럭킹 등으로 인해 CPU가 소음과 발열이 심해져서 자리가 옮겨졌는데, 이 경우 보증 수리가 가능한가요? |

# Appendix B  Code-mixed sample items

**Table 25.** The first code-mixed test item of every variety for the billing and technical departments.

| Variety | Gold | Message |
|---|---|---|
| Hinglish | billing | refund status update karo, bank statement me kuch nahi dikha raha |
| Hinglish | technical | bhai login page pe gir gaya hai, refresh karne se bhi kaam nahi karta, koi bug aa gaya kya? |
| Spanglish | billing | Mi tarjeta fue rechazada al intentar hacer el payment de la suscripción premium, pero me cobraron los $49.99 de todos modos, necesito un refund urgent… |
| Spanglish | technical | No puedo verificar 2FA porque el SMS […] está down, los usuarios no reciben el OTP code, urgente fix por favor. |
| Arabizi | billing | zayd auto-debit kharra, lakan I cancelled sub 3 days ago, lakan charge still hit my card? please reverse this txn immediately, lakan customer service… |
| Arabizi | technical | Random crashes on mobile app after update to v2.4, stack trace shows native crash in swift code, need dev team to investigate core dump |

# Appendix C  Files and checksums

**Table 26.** Data files with row counts and SHA-256 (first 20 hex digits).

| File | Rows | SHA-256 |
|---|---:|---|
| `data/ml_test.jsonl` | 322 | `d3d9d5e7c21febc477e0…` |
| `data/mix_test.jsonl` | 156 | `52d0aa54d37e1a3aea80…` |
| `data/ml_train.jsonl` | 967 | `2d4b0c4230fbe3466ec7…` |
| `data/train.jsonl` | 2818 | `18a3c345b94ac8e3830c…` |
| `data/train_ml.jsonl` | 3785 | `b64cad60df11156ca1fe…` |
| `data/train_ml_ho.jsonl` | 3464 | `342a71678c3dc8d3986f…` |
| `results/baselines.json` | – | `2cd5bf98e456091f629b…` |
| `results/xlabel_gemma4-31b.json` | – | `28f1fa32695a39f63cc4…` |

# Appendix D  Prompts and scoring recipe

## D.1  Generation request (paraphrased)

> Write *k* (at most 10) diverse, realistic inputs for a classifier. Question: "Which department should handle this?" The correct answer for every input must be <*department*> (with its one-line description). Write every text in <*language*> only (native script), with no English words except product names. Make them unambiguous to a careful human and vary length, style and vocabulary (variation seed *s*). Return only a JSON array of strings.

## D.2  Blind relabelling request (paraphrased)

> Question and the three options with their descriptions are shown, followed by the numbered texts of one call; for each text choose exactly one option key; return only a JSON array of keys in order. The intended label is not shown. A text is kept if and only if the returned key equals the intended label.

## D.3  Scoring recipe

1. Load `data/ml_test.jsonl` and `data/mix_test.jsonl`; each row has `state` (the message), `questions` (the typed question) and `expected` (the gold option).
2. For every row build the input exactly as published and obtain one probability per option in the order billing, technical, other.
3. Prediction = arg-max; confidence = its probability; correct = prediction equals the gold option.
4. Report accuracy with a Wilson interval, language-macro accuracy, worst language, macro-F1, NLL and ECE, on the native and code-mixed splits separately, and the track under which the system was trained.